\documentclass{midl} 

\usepackage{graphicx}
\usepackage{mathabx}
\usepackage{booktabs}

\jmlrvolume{-- Under Review}
\jmlryear{2025}
\jmlrworkshop{Full Paper -- MIDL 2025 submission}
\editors{Under Review for MIDL 2025}

\title[Single-Step Denoising Diffusion-GAN (SSDD-GAN)]{SSDD-GAN: Single-Step Denoising Diffusion GAN for Cochlear Implant Surgical Scene Completion}

\midlauthor{\Name{Yike Zhang\nametag{$^{1}$}} \Email{yike.zhang@vanderbilt.edu}\\
\Name{Eduardo Davalos\nametag{$^{2}$}} \Email{eduardo.davalos.anaya@vanderbilt.edu}\\
\Name{Jack Noble\nametag{$^{3}$}} \Email{jack.noble@vanderbilt.edu}
}

\begin{document}

\maketitle

\begin{abstract}
Recent deep learning-based image completion methods, including both inpainting and outpainting, have demonstrated promising results in restoring corrupted images by effectively filling various missing regions. Among these, Generative Adversarial Networks (GANs) and Denoising Diffusion Probabilistic Models (DDPMs) have been employed as key generative image completion approaches, excelling in the field of generating high-quality restorations with reduced artifacts and improved fine details. In previous work, we developed a method aimed at synthesizing views from novel microscope positions for mastoidectomy surgeries; however, that approach did not have the ability to restore the surrounding surgical scene environment.
In this paper, we propose an efficient method to complete the surgical scene of the synthetic postmastoidectomy dataset. Our approach leverages self-supervised learning on real surgical datasets to train a Single-Step Denoising Diffusion-GAN (SSDD-GAN), combining the advantages of diffusion models with the adversarial optimization of GANs for improved Structural Similarity results of 6\%. The trained model is then directly applied to the synthetic postmastoidectomy dataset using a zero-shot approach, enabling the generation of realistic and complete surgical scenes without the need for explicit ground-truth labels from the synthetic postmastoidectomy dataset. This method addresses key limitations in previous work, offering a novel pathway for full surgical microscopy scene completion and enhancing the usability of the synthetic postmastoidectomy dataset in surgical preoperative planning and intraoperative navigation.
\end{abstract}

\begin{keywords}
Image completion, image inpainting, image outpainting, image synthesis, surgical scene synthesis, postmastoidectomy, cochlear implant surgery, Denoising Diffusion Probabilistic Models (DDPMs), Generative Adversarial Networks (GANs), Diffusion-GAN.
\end{keywords}

\section{Introduction}
Cochlear Implant (CI) procedures are transformative surgeries that aim to restore hearing for individuals with moderate-to-profound hearing disabilities, offering a way to improve communication and quality of life \cite{labadie2018preliminary}. These procedures involve the precise placement of an electrode array into the cochlea, enabling direct stimulation of the auditory nerve to restore hearing ability \cite{10.1117/12.2655653, 10.1117/12.3008830}. As one of the initial steps in CI surgery, mastoidectomy involves the careful removal of portions of the temporal bone to create access to the middle ear and cochlea. This procedure ensures a clear pathway for electrode array insertion while safeguarding critical anatomical structures, such as the facial nerve and the chorda. We hypothesize that if the surgically created mastoidectomy surface can be predicted directly from preoperative CT scans, it could serves as a valuable resource for numerous downstream tasks, including surgical tool tracking, surgical scene synthesis, and pose estimation of anatomical structures. These potential benefits could collectively contribute to improved surgical navigation and enhanced intraoperative visualization, ultimately supporting greater precision and optimizing the placement of the electrode array during cochlear implantation.

Recent studies have increasingly focused on leveraging advanced imaging and deep learning-based methods to assist surgeons in understanding and navigating complex anatomical structures during cochlear implant surgery. In our previous work \cite{zhang2024mmunsupervisedmambabasedmastoidectomy, zhang2024mastoidectomymultiviewsynthesissingle}, we developed novel methodologies to reconstruct the postmastoidectomy surface and novel views from a single microscopy image. These methods demonstrated significant potential in providing partial reconstructions of the surgical scene, improving intraoperative visualization, and eliminating reliance on external tracking devices. However, these approaches were limited to texturing the postmastoidectomy surface from preoperative CT scans, neglecting the broader surgical environment captured by the microscopy.
The absence of contextual information surrounding the surgical site poses challenges for comprehensive scene understanding, particularly in scenarios where broader spatial awareness is critical for decision-making. To address these limitations, this paper proposes a novel approach that leverages image completion using a deep learning-based generative model to fill in the missing regions of the surgical scene, enabling the synthesis of a complete and detailed surgical environment.
Image completion involves reconstructing missing or occluded regions (inpainting) or extending an image beyond its boundaries (outpainting) by leveraging contextual information from the known areas. It is a critical task in computer vision with wide-ranging applications in photo editing, image-based rendering, and computational photography \cite{park2017transformationgroundedimagegenerationnetwork, sabini2018paintingoutsideboximage, DBLP:journals/corr/abs-2109-07161}. The primary challenge lies in generating visually realistic and semantically meaningful pixels for the missing regions while ensuring seamless coherence with the known content. 

Recent popular works in image inpainting and outpainting are heavily based on deep learning neural networks. In the research proposed by \cite{7780647}, a context encoder was introduced to predict missing image regions using convolutional neural networks (CNNs), laying the foundation for generative approaches to image inpainting. Subsequent advancements, such as those by \cite{10.1145/3072959.3073659}, incorporated both global and local discriminators to enhance texture consistency, while \cite{yu2018generativeimageinpaintingcontextual} introduced the use of contextual attention mechanisms for more realistic inpainting of irregular holes. These developments have significantly improved the quality and applicability of image completion techniques across various domains. 
Surgical data often contains complex anatomical structures, cluttered scenes, and occlusions caused by surgical tools or the surgeon's hands. Beyond visual realism, inpainting for surgical scenes requires clinically meaningful reconstructions with high geometric fidelity to preserve critical anatomical details. These challenges require novel and effective approaches that can adapt to the complexities of real surgical scenes without relying heavily on manual annotations. For instance, \cite{DAHER2023102994} introduced a machine learning approach using a temporal generative adversarial network (GAN) to inpaint hidden anatomy under specularities.
This paper aims to reconstruct the complete post-mastoidectomy surgical scene by training a neural network on real surgical datasets. Additionally, it focuses on building a patient-specific dataset to assist intraoperative registration between preoperative CT scans and the corresponding surgical scene. With the proposed self-supervised Single-Step Denoising Diffusion-GAN (SSDD-GAN) framework, our approach bypasses the need for manual annotations by learning directly from the inherent structures in the data, enabling accurate and clinically relevant surgical scene synthesis.
Our contributions can be summarized in the following:
\begin{itemize}
    \item \textbf{Novel Self-supervised Image Completion Framework SSDD-GAN:} We introduce an image completion framework SSDD-GAN that aims for surgical scene inpainting and outpainting. This self-supervised approach eliminates the need for manually annotated datasets, ensuring training efficiency and generalizability.
    \item \textbf{Zero-shot Synthesis using the Synthetic Postmastoidectomy Dataset:} Our goal is to generate a complete surgical scene by utilizing the synthetic postmastoidectomy dataset via a zero-shot learning strategy by training and validating the model on real surgical datasets.
    \item \textbf{Enhanced Cochlear Implant Surgery Visualization and Navigation:} The proposed method provides full surgical field visualizations along with precise camera pose information derived from the previously synthetic postmastoidectomy dataset. This advancement paves the way for surgical scene understanding, tool tracking, and anatomical navigation, offering the potential for improving cochlear implant surgery preoperative planning and intraoperative guidance.
\end{itemize}

\section{Methodology}
\begin{figure}[ht]
    \centering
    \includegraphics[width=0.9\textwidth]{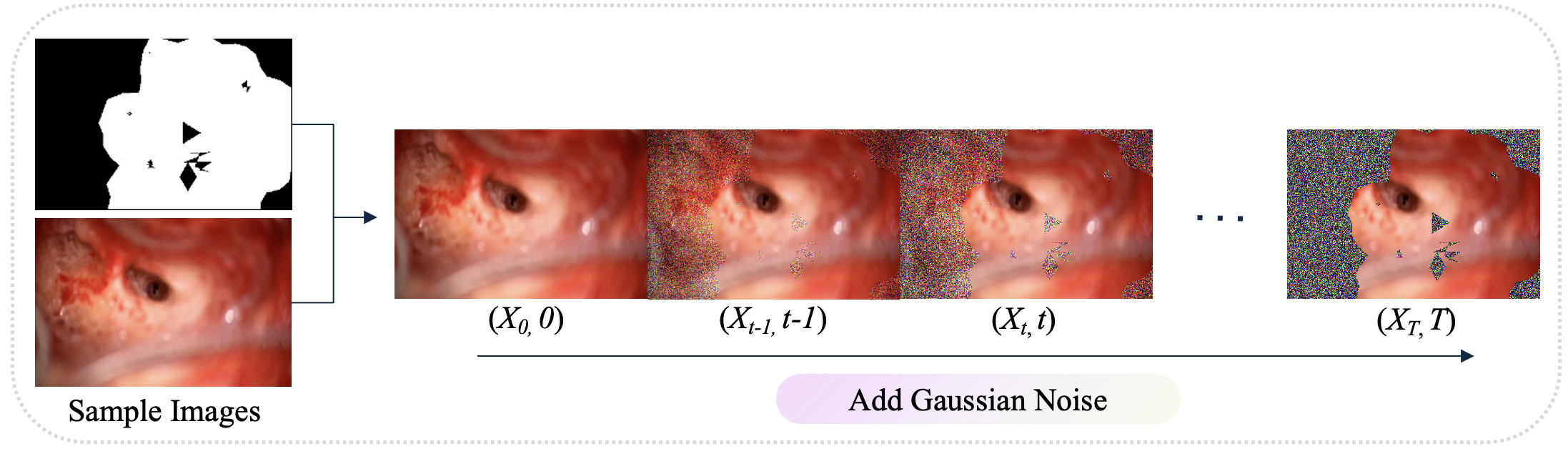}
    \caption{\textbf{Forward Diffusion Process}. We preserve the masked region of the original sample data while applying Gaussian noise exclusively to the non-masked region.}
    \label{fig:diffusion_process}
\end{figure}

To address the limitations outlined in \cite{zhang2024mastoidectomymultiviewsynthesissingle} and enable the completion of a surgical scene, we propose a deep-learning-based approach trained and validated on a dataset of real microscopy views. Given the irregular shapes of the synthetic postmastoidectomy surgical views (shown in the first row of Figure. \ref{fig:synthesis}), we generate random masks on the real surgical dataset to simulate the partially generated postmastoidectomy multi-views. This dataset creation strategy ensures that the model effectively learns to restore missing regions while maintaining robustness to the diverse shapes and irregularities of the synthetic postmastoidectomy scenes. 
To simulate the partially generated surgical scenes using the postmastoidectomy surface, we automatically generate masks on the surgical frames using a range of polygonal shapes containing randomly placed holes. This label-generation approach effectively mimics the irregularities and variability observed in the synthetic post-mastoidectomy scene dataset. 
We propose SSDD-GAN that combines the strengths of diffusion models and GANs to synthesize realistic surgical scenes guided by randomly masked real surgical data. While traditional diffusion models have noticeable advantages in generating synthetic images, audio, and videos, they often suffer from slow inference times due to their long iterative sampling process. This limitation also presents challenges when attempting to integrate a discriminator into the denoising routine. 
Unlike traditional DDPMs \cite{ho2020denoisingdiffusionprobabilisticmodels}, which rely on the iterative denoising process, our method focuses exclusively on single-step denoising and reconstruction to minimize computational cost by directly mapping noise to data. Adding a discriminator to the single-step denoising routine can lead to satisfactory results since the discriminator provides an additional adversarial learning signal that helps the diffusion model refine the predicted noises and leads to better outputs.
Leveraging this unique feature, we aim to enhance sampling efficiency and seamlessly integrate a discriminator into the single-step denoising process, further improving the quality and realism of the generated surgical scenes.
The forward diffusion process of our method is shown in Figure~\ref{fig:diffusion_process}. As shown in the figure, we only apply the Gaussian noise on the non-masked region in the forward diffusion process.
The data points for the forward diffusion process are sampled from a real data distribution $x_t \sim q(x)$. This process progressively adds Gaussian noise to the targeted region in the samples over $T$ steps, where $T \in [700, 900]$, as determined by our experiments. We produce a sequence of noisy samples $x_1, ..., x_T$. The interval sizes are controlled by a linear beta scheduler $\{\beta_t \in (0, 1)\}_{t=1}^T$. The whole standard forward diffusion process can be expressed in the following Eq~\ref{eq:forward_diffusion}:
\begin{equation}
\begin{split}
    q(x_t|x_{t-1}) = \mathcal{N}(x_{t}; \sqrt{1 - \beta_t}x_{t-1}, (\sqrt{\beta_t})^2 I), \\
    q(x_t|x_{t-1}) = \mathcal{N}(x_{t}; \sqrt{1 - \beta_t}x_{t-1}, \beta_t I), \\
    q(x_{1:T}|x_0) = \prod_{t=1}^{T}q(x_t|x_{t-1})
\end{split}
\label{eq:forward_diffusion}
\end{equation}
At any arbitrary time step $t$, we can sample $x_t$ in a closed form using the re-parametrization method, which the method can be described as the following Eq~\ref{eq:reparametrization}. We set the random variable $z$, $q_{\phi}(z|x)$ as a multivariate Gaussian, and $\epsilon$ is an auxiliary independent random variable.
\begin{equation}
\begin{split}
z \sim q_\phi(z|x^{i}) = \mathcal{N}(z; \mu^{i}, \sigma^{2(i)} I), \\
z = \mu + \sigma \odot \epsilon,\; \text{where } \epsilon \sim \mathcal{N}(0, I)    
\end{split}
\label{eq:reparametrization}
\end{equation}
Gaussian noise can be directly added from $x_0$ to any arbitrary step $x_t$ using the following Eq~\ref{eq:add_noise}. Let $\alpha_t = 1 - \beta_t$, $\beta_t = 1 - \alpha_t$, $\widebar{\alpha_t} = \prod_{i=1}^{t}\alpha_{i}$, and $\delta$ denotes for the generated mask regions:
\begin{equation}
\begin{aligned}
    x_t &= (\sqrt{\alpha_t}x_{t-1} + \sqrt{1 - \alpha_t}\epsilon_{t-1})(1 - \delta) + \delta x_0, \quad \text{where} \; \epsilon_{t-1} \sim \mathcal{N}(0, I), \\
    &=(\sqrt{\alpha_t\alpha_{t-1}}x_{t-2} + \sqrt{1 - \alpha_t\alpha_{t-1}}\epsilon_{t-1}\epsilon_{t-2})(1 - \delta) + \delta x_0, \\
    &= ... \\
    &=(\sqrt{\bar{\alpha_t}}x_0 + \sqrt{1 - \bar{\alpha_t}}\bar{\epsilon}) (1 - \delta) + \delta x_0, \quad \text{where} \; \bar{\epsilon} \text{ merges $\epsilon_{t-1}$, $\epsilon_{t-2}$, ... Gaussians}.
\end{aligned}
\label{eq:add_noise}
\end{equation}
The term $(\sqrt{\bar{\alpha_t}}x_0 + \sqrt{1 - \bar{\alpha_t}}\bar{\epsilon}) (1 - \delta)$ represents a partial forward-diffusion mix of the clean image $x_0$ and merged Gaussian noise $\bar{\epsilon}$, scaled by $(1 - \delta)$. The term $\delta x_0$ adds back a fraction $\delta$ of the original image $x_0$ to $x_t$. We progressively reduce the signal in the masked region of the original image sample $x_0$ by a factor of $\sqrt{\bar{\alpha_t}}$, while simultaneously adding noise to the masked region scaled by $\sqrt{1 - \bar{\alpha_t}}\bar{\epsilon}$. 
\begin{figure}[ht]
    \centering
    \includegraphics[width=0.9\textwidth]{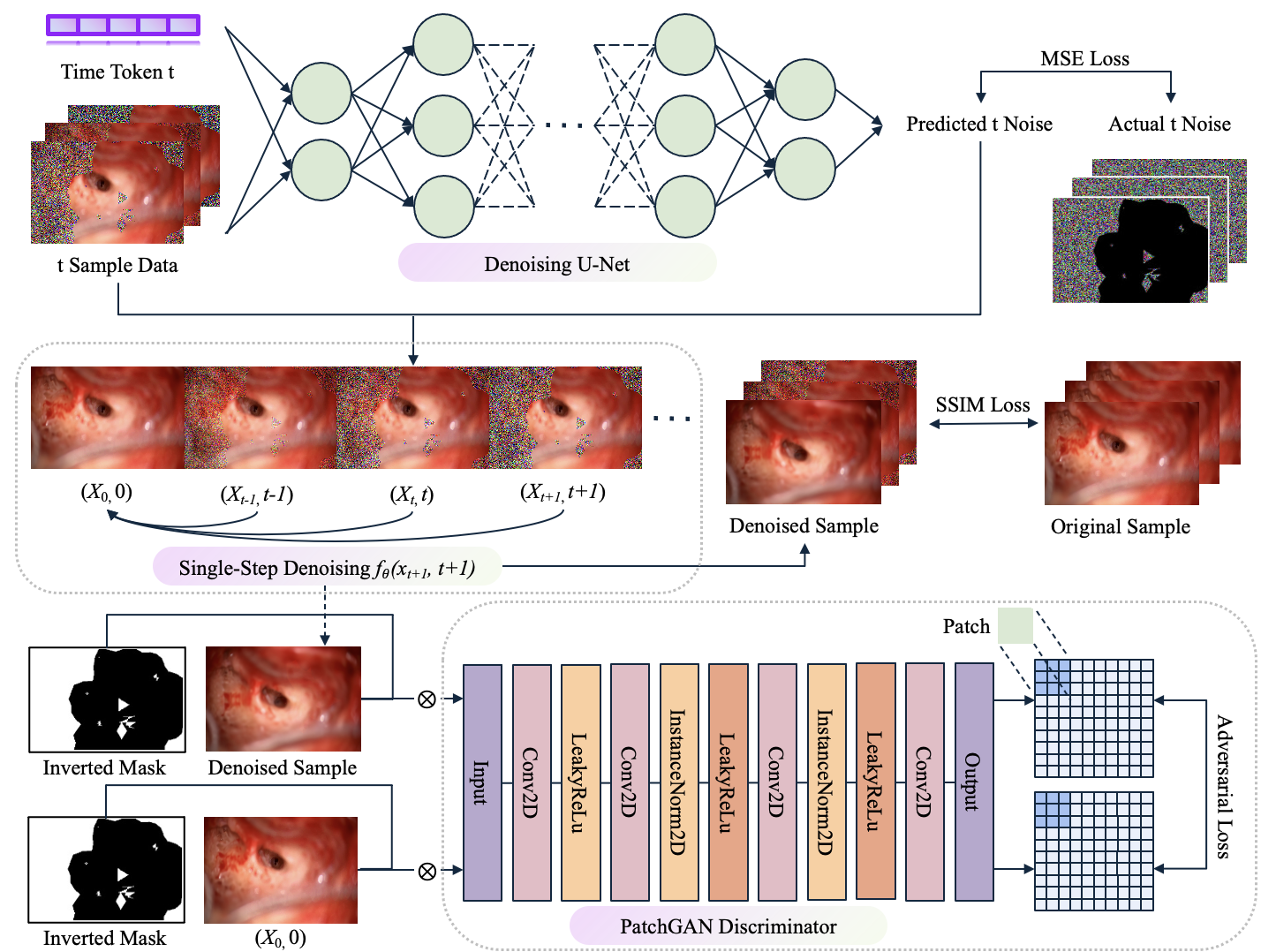}
    \caption{\textbf{Single-Step Denoising Diffusion Process}. We incorporate a discriminator in this process to further improve the realism of synthetic samples.}
    \label{fig:ssdd}
\end{figure}
The proposed single-step denoising process is shown in Figure~\ref{fig:ssdd}. The denoising U-Net structure is adopted from the method proposed in \cite{ho2020denoisingdiffusionprobabilisticmodels}. Unlike the iterative denoising process proposed in their method, our approach directly employs a neural network to predict the noise $\epsilon_t$ and map any arbitrary step $x_t$ to a reconstruction of $x_0$ in a single step, shown in Eq~\ref{eq:denoising_process}:
\begin{equation}
x_0 = \left(\frac{x_t - \sqrt{1 - \bar{\alpha_t}}\bar{\epsilon}}{\sqrt{\bar{\alpha_t}}}\right)(1 - \delta) + \delta x_t
\label{eq:denoising_process}
\end{equation}
Note that $\delta x_t$ is equivalent to $\delta x_0$ in our setting. During training, we use Mean Squared Error (MSE) loss for noise prediction. The reconstructed images are then compared with their corresponding sample data using Structural Similarity Index (SSIM) loss, further refining the model and improving the reconstruction quality. This direct denoising method significantly reduces computational time while enabling the integration of a discriminator for enhanced performance. Specifically, we implement a Patch-GAN discriminator\cite{isola2018imagetoimagetranslationconditionaladversarial}, which evaluates image structure at the patch level. The discriminator classifies whether each $N$ by $N$ patch in an image is real or fake by applying a convolutional filter across the entire image and gathering the responses to produce the final output. This approach ensures that the model focuses on local details while maintaining computational efficiency. The discriminator is trained using the BCEWithLogits loss function that focuses solely on inputs from the generated content region and the corresponding real surgical scene region.
\section{Results}
\label{sec:results}
\begin{figure}[ht]
  \centering
  \begin{minipage}{0.32\textwidth}
        \centering
        \includegraphics[width=\textwidth]{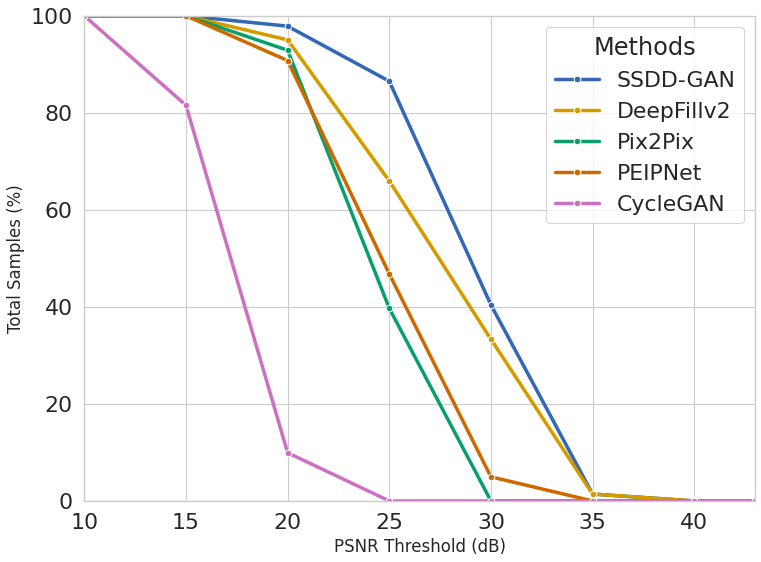}
        \label{fig:psnr_lines}
        \footnotesize{(a) PSNR comparison}
    \end{minipage}
    \hfill
    \begin{minipage}{0.32\textwidth}
        \centering
        \includegraphics[width=\textwidth]{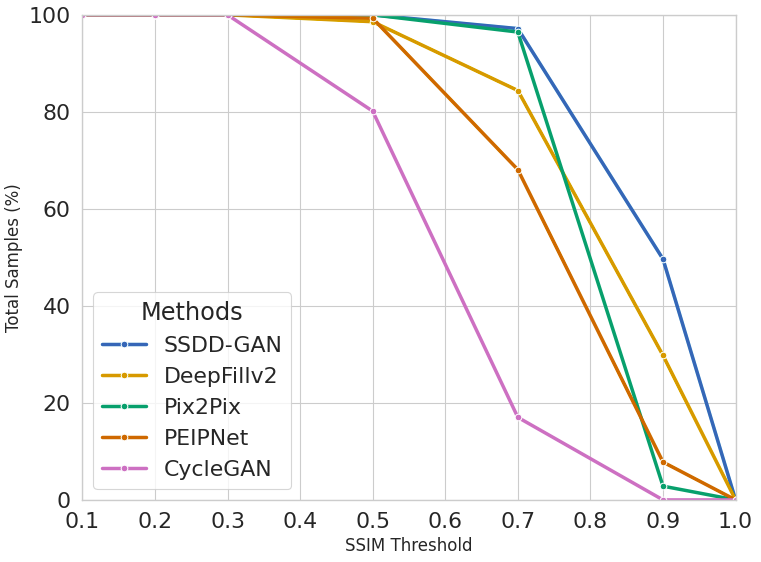} 
        \label{fig:ssim_lines}
        \footnotesize{(b) SSIM comparison}
    \end{minipage}
    \hfill
    \begin{minipage}{0.32\textwidth}
        \centering
        \includegraphics[width=\textwidth]{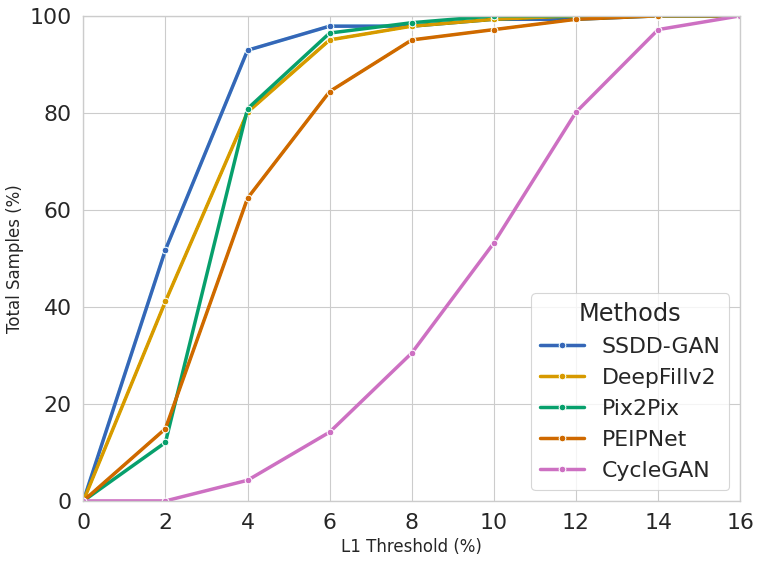}
        \label{fig:MAPE_Lines}
        \footnotesize{(c) L1(\%) comparison}
    \end{minipage}
    \begin{minipage}{0.32\textwidth}
        \centering
        \includegraphics[width=\textwidth]{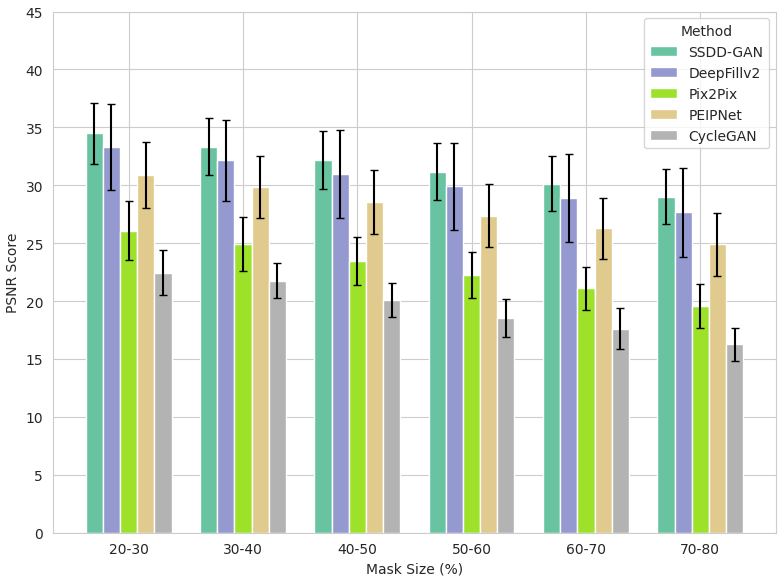} 
        \label{fig:PSNR_bar}
        \footnotesize{(d) PSNR Comparison Across Varying Mask Ratios}
    \end{minipage}
    \hfill
    \begin{minipage}{0.32\textwidth}
        \centering
        \includegraphics[width=\textwidth]{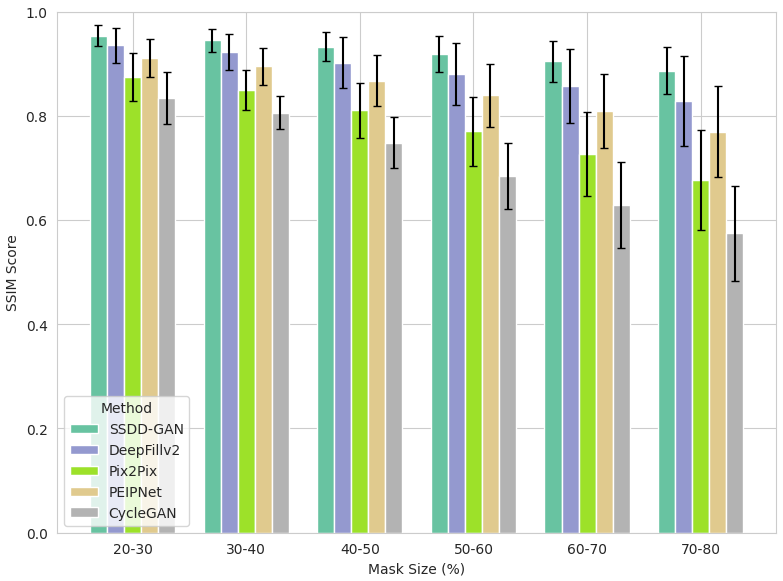}
        \label{fig:SSIM_bar}
        \footnotesize{(e) SSIM Comparison Across Varying Mask Ratios}
    \end{minipage}
    \hfill
    \begin{minipage}{0.32\textwidth}
        \centering
        \includegraphics[width=\textwidth]{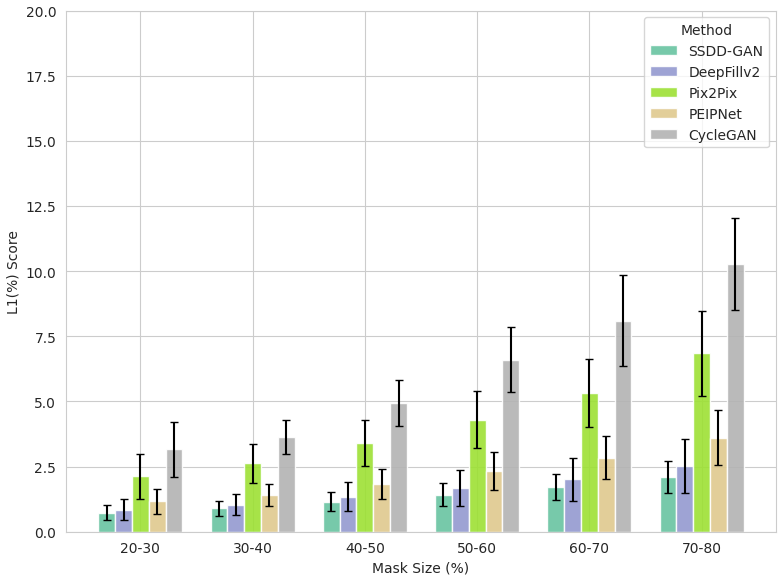}
        \label{fig:MAPE_bar}
        \footnotesize{(f) L1(\%) Comparison Across Varying Mask Ratios}
    \end{minipage}
\caption{\textbf{Performance Comparisons}. The experiments evaluate overall performance (\textbf{top row}) as well as performance across varying mask ratios \textbf{(bottom row)}.}
\label{fig:performance_overall_comparison}
\end{figure}

\begin{figure}[ht]
  \centering
  \begin{minipage}{0.136\textwidth}
        \includegraphics[width=\textwidth]{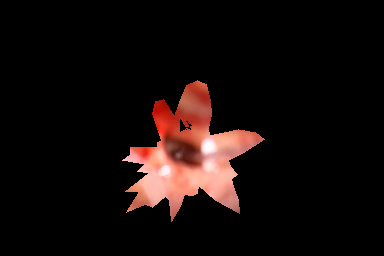}
    \end{minipage}
    \begin{minipage}{0.136\textwidth}
        \includegraphics[width=\textwidth]{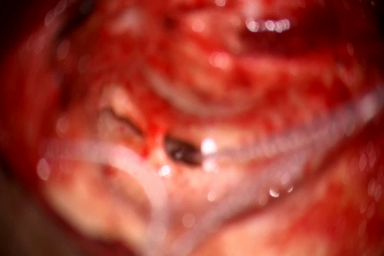}
    \end{minipage}
    \begin{minipage}{0.136\textwidth}
        \includegraphics[width=\textwidth]{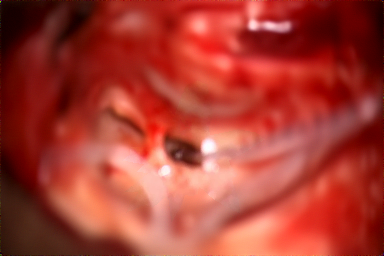}
    \end{minipage}
    \begin{minipage}{0.136\textwidth}
        \includegraphics[width=\textwidth]{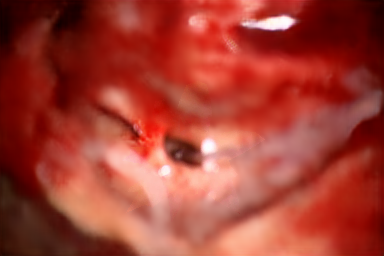}
    \end{minipage}
    \begin{minipage}{0.136\textwidth}
        \includegraphics[width=\textwidth]{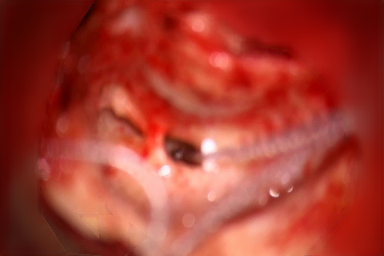}
    \end{minipage}
    \begin{minipage}{0.136\textwidth}
        \includegraphics[width=\textwidth]{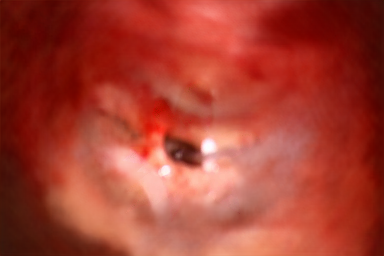}
    \end{minipage}
    \begin{minipage}{0.136\textwidth}
        \includegraphics[width=\textwidth]{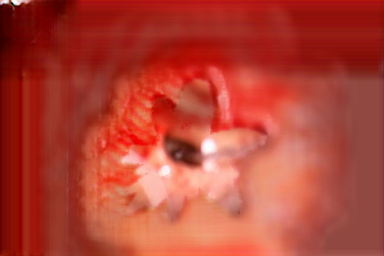}
    \end{minipage}
    \begin{minipage}{0.136\textwidth}
        \includegraphics[width=\textwidth]{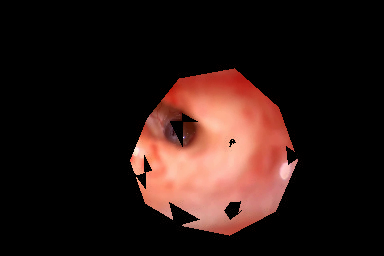}
    \end{minipage}
    \begin{minipage}{0.136\textwidth}
        \includegraphics[width=\textwidth]{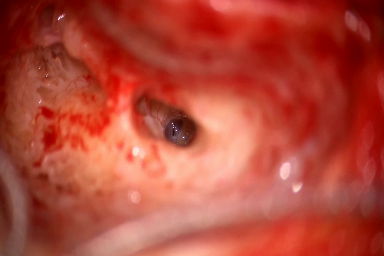}
    \end{minipage}
    \begin{minipage}{0.136\textwidth}
        \includegraphics[width=\textwidth]{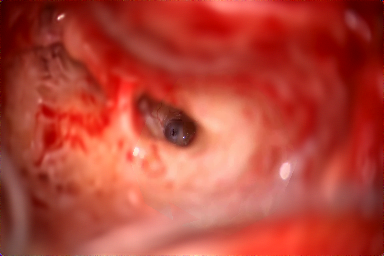}
    \end{minipage}
    \begin{minipage}{0.136\textwidth}
        \includegraphics[width=\textwidth]{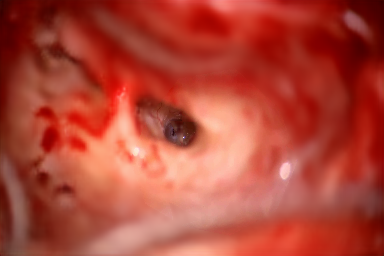}
    \end{minipage}
    \begin{minipage}{0.136\textwidth}
        \includegraphics[width=\textwidth]{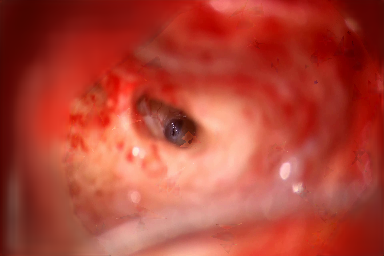}
    \end{minipage}
    \begin{minipage}{0.136\textwidth}
        \includegraphics[width=\textwidth]{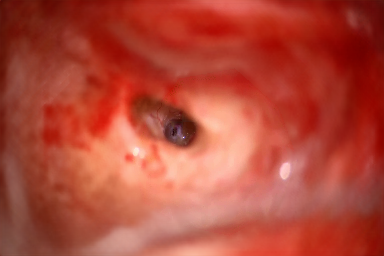}
    \end{minipage}
    \begin{minipage}{0.136\textwidth}
        \includegraphics[width=\textwidth]{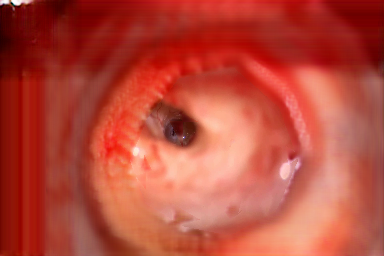}
    \end{minipage}
    \begin{minipage}{0.136\textwidth}
        \includegraphics[width=\textwidth]{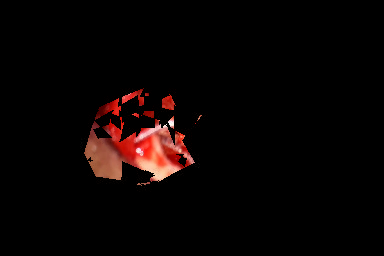}
    \end{minipage}
    \begin{minipage}{0.136\textwidth}
        \includegraphics[width=\textwidth]{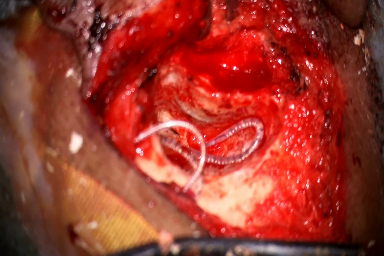}
    \end{minipage}
    \begin{minipage}{0.136\textwidth}
        \includegraphics[width=\textwidth]{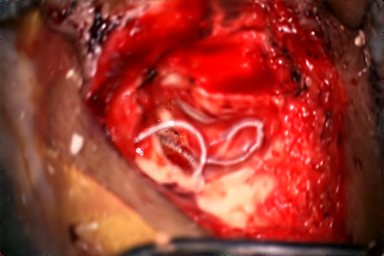}
    \end{minipage}
    \begin{minipage}{0.136\textwidth}
        \includegraphics[width=\textwidth]{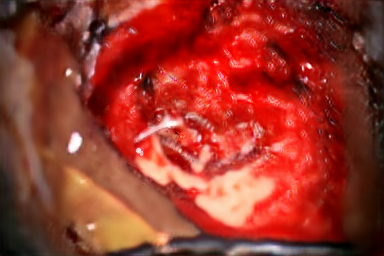}
    \end{minipage}
    \begin{minipage}{0.136\textwidth}
        \includegraphics[width=\textwidth]{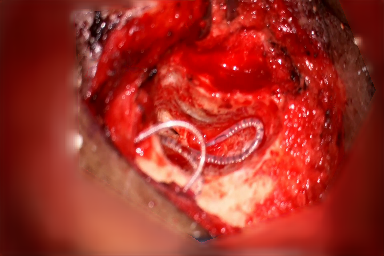}
    \end{minipage}
    \begin{minipage}{0.136\textwidth}
        \includegraphics[width=\textwidth]{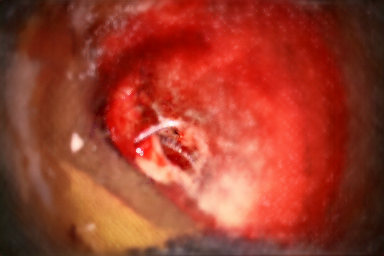}
    \end{minipage}
    \begin{minipage}{0.136\textwidth}
        \includegraphics[width=\textwidth]{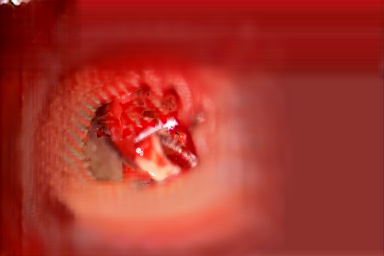}
    \end{minipage}
    \begin{minipage}{0.136\textwidth}
        \centering
        \includegraphics[width=\textwidth]{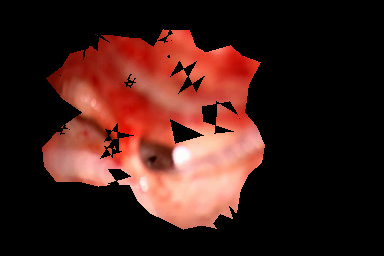}
        \footnotesize{Input}
    \end{minipage}
    \begin{minipage}{0.136\textwidth}
        \centering
        \includegraphics[width=\textwidth]{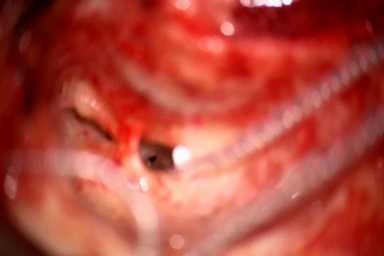}
        \footnotesize{GT}
    \end{minipage}
    \begin{minipage}{0.136\textwidth}
        \centering
        \includegraphics[width=\textwidth]{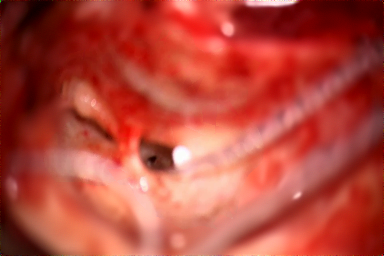}
        \footnotesize{SSDD-GAN}
    \end{minipage}
    \begin{minipage}{0.136\textwidth}
        \centering
        \includegraphics[width=\textwidth]{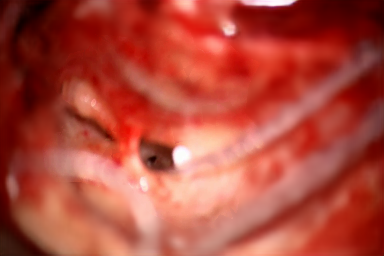}
        \footnotesize{DeepFillv2}
    \end{minipage}
    \begin{minipage}{0.136\textwidth}
    \centering
        \includegraphics[width=\textwidth]{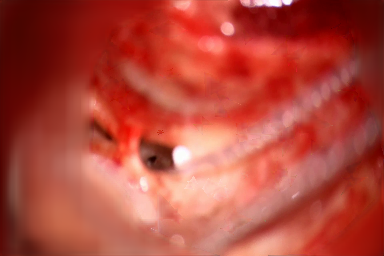}
        \footnotesize{Pix2Pix}
    \end{minipage}
    \begin{minipage}{0.136\textwidth}
    \centering
        \includegraphics[width=\textwidth]{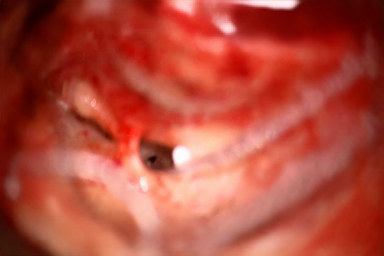}
        \footnotesize{PEIPNet}
    \end{minipage}
    \begin{minipage}{0.136\textwidth}
    \centering
        \includegraphics[width=\textwidth]{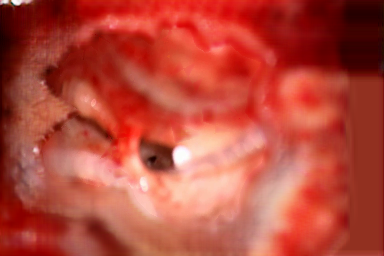}
        \footnotesize{CycleGAN}
    \end{minipage}
\caption{\textbf{Qualitative Comparisons}. Visualizations of completing missing regions using various methods. Certain details are highlighted in cyan bounding boxes.}
\label{fig:representative_samples}
\end{figure}

\begin{figure}[ht]
  \centering
  \begin{minipage}{0.32\textwidth}
        \centering
        \includegraphics[width=\textwidth]{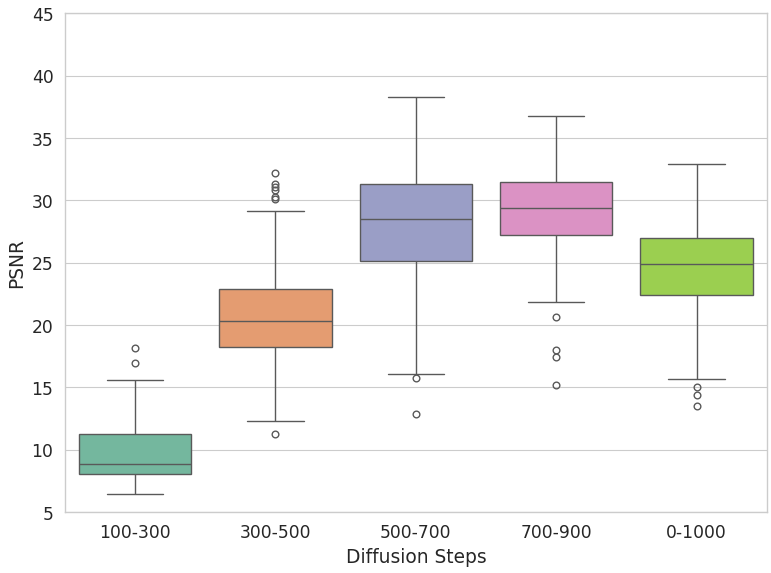}
        \footnotesize{(a) PSNR comparison}
    \end{minipage}
    \hfill
    \begin{minipage}{0.32\textwidth}
        \centering
        \includegraphics[width=\textwidth]{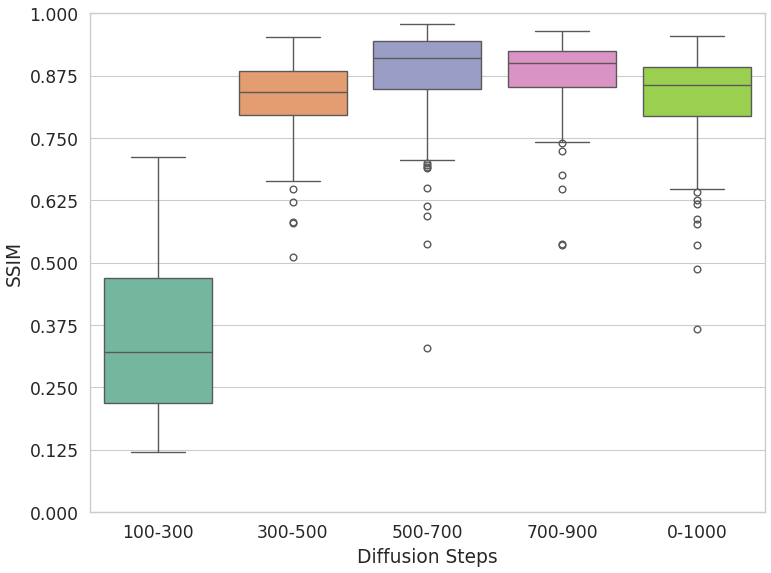}
        \footnotesize{(b) SSIM comparison}
    \end{minipage}
    \hfill
    \begin{minipage}{0.32\textwidth}
        \centering
        \includegraphics[width=\textwidth]{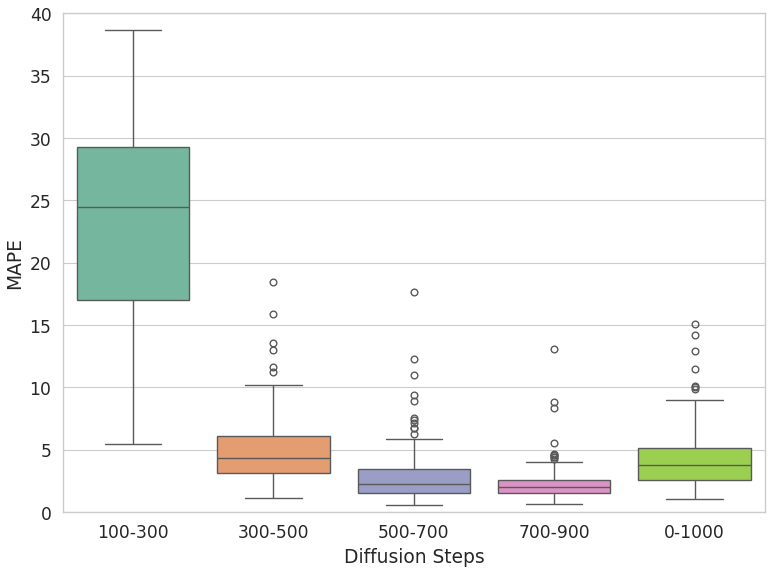}
        \footnotesize{(c) L1(\%) comparison}
    \end{minipage}
\caption{\textbf{Ablation Study}. Analyzing the impact of varying the number of $T$.}
\label{fig:ablation_study}
\end{figure}

\begin{figure}[h!]
  \centering
  \begin{minipage}{0.16\textwidth}
        \includegraphics[width=\textwidth]{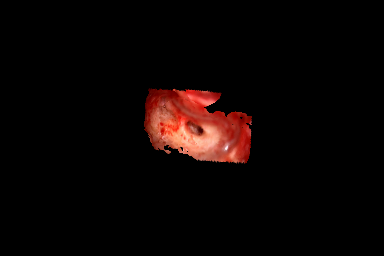}
    \end{minipage}
    \hfill
    \begin{minipage}{0.16\textwidth}
        \includegraphics[width=\textwidth]{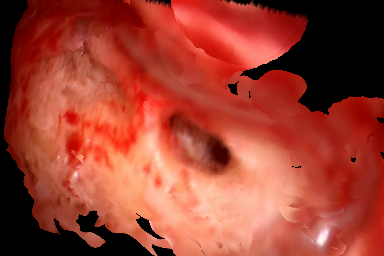}
    \end{minipage}
    \hfill
    \begin{minipage}{0.16\textwidth}
        \includegraphics[width=\textwidth]{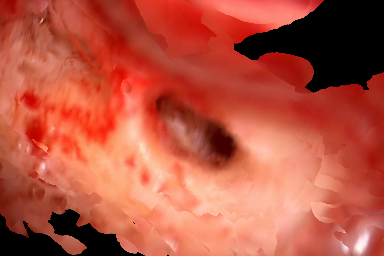}
    \end{minipage}
    \hfill
    \begin{minipage}{0.16\textwidth}
        \includegraphics[width=\textwidth]{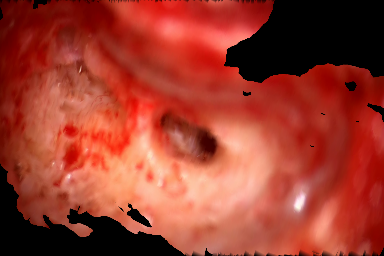}
    \end{minipage}
    \hfill
    \begin{minipage}{0.16\textwidth}
        \includegraphics[width=\textwidth]{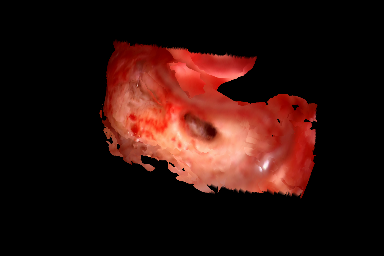}
    \end{minipage}
    \hfill
    \begin{minipage}{0.16\textwidth}
        \includegraphics[width=\textwidth]{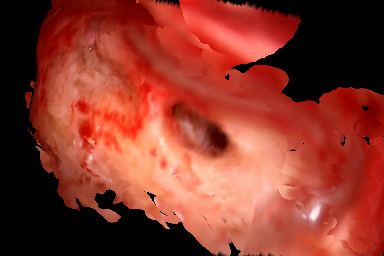}
    \end{minipage}
    \begin{minipage}{0.16\textwidth}
        \includegraphics[width=\textwidth]{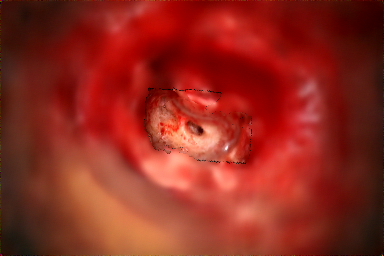}
    \end{minipage}
    \hfill
    \begin{minipage}{0.16\textwidth}
        \includegraphics[width=\textwidth]{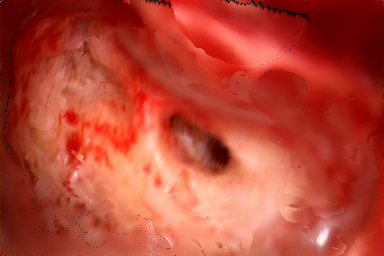}
    \end{minipage}
    \hfill
    \begin{minipage}{0.16\textwidth}
        \includegraphics[width=\textwidth]{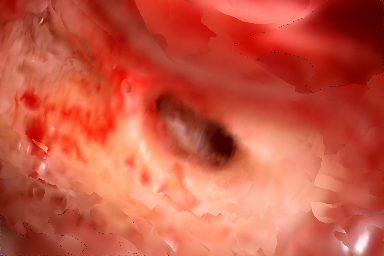}
    \end{minipage}
    \hfill
    \begin{minipage}{0.16\textwidth}
        \includegraphics[width=\textwidth]{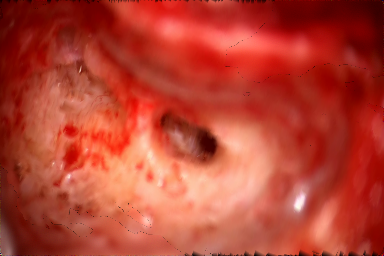}
    \end{minipage}
    \hfill
    \begin{minipage}{0.16\textwidth}
        \includegraphics[width=\textwidth]{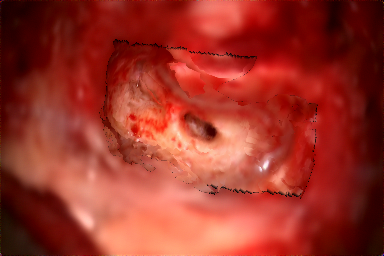}
    \end{minipage}
    \hfill
    \begin{minipage}{0.16\textwidth}
        \includegraphics[width=\textwidth]{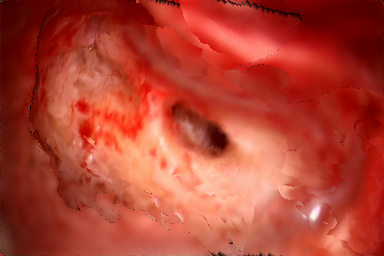}
    \end{minipage}
    \hfill
    \begin{minipage}{0.16\textwidth}
        \centering
        \includegraphics[width=\textwidth]{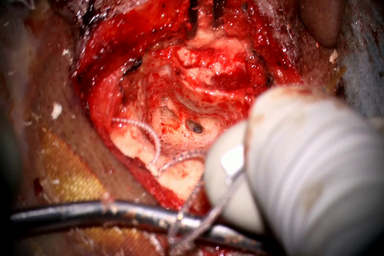}
        \footnotesize{Sample 1}
    \end{minipage}
    \begin{minipage}{0.16\textwidth}
        \centering
        \includegraphics[width=\textwidth]{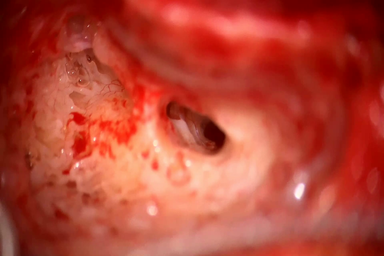}
        \footnotesize{Sample 2}
    \end{minipage}
    \hfill
    \begin{minipage}{0.16\textwidth}
        \centering
        \includegraphics[width=\textwidth]{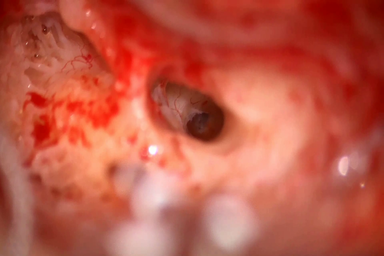}
        \footnotesize{Sample 3}
    \end{minipage}
    \hfill
    \begin{minipage}{0.16\textwidth}
        \centering
        \includegraphics[width=\textwidth]{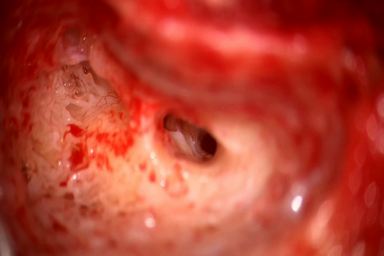}
        \footnotesize{Sample 4}
    \end{minipage}
    \hfill
    \begin{minipage}{0.16\textwidth}
        \centering
        \includegraphics[width=\textwidth]{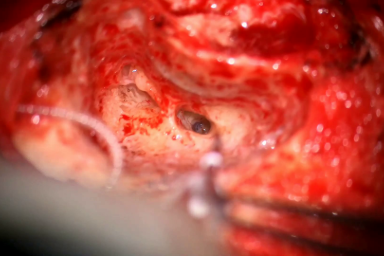}
        \footnotesize{Sample 5}
    \end{minipage}
    \hfill
    \begin{minipage}{0.16\textwidth}
        \centering
        \includegraphics[width=\textwidth]{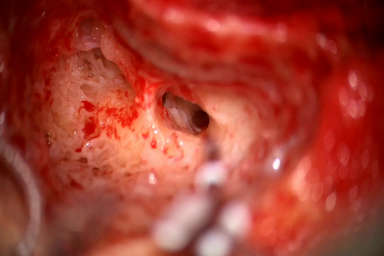}
        \footnotesize{Sample 6}
    \end{minipage}
\caption{\textbf{Surgical Scene Synthesis}. Results of reconstructing the complete surgical field using the synthetic postmastoidectomy dataset. The first row shows original data, the second row presents the completed surgical scenes, and the final row displays the closest corresponding real surgical scenes.}
\label{fig:synthesis}
\end{figure}

Our dataset comprises 932 real surgical frames collected from a cochlear implant surgery on a patient, divided into training, validation, and testing sets in a ratio of 0.75, 0.15, and 0.15, respectively.
The quantitative results are summarized in Table~\ref{Tab:quantitative}, comparing our method to other generative models, such as CycleGAN\cite{zhu2020unpairedimagetoimagetranslationusing}, Pix2Pix\cite{isola2018imagetoimagetranslationconditionaladversarial}, DeepFillv2\cite{yu2019freeformimageinpaintinggated}, and PEIPNet\cite{PEIPNet}. We use metrics such as Fréchet Inception Distance (FID)\cite{fid}, Kernel Inception Distance (KID)\cite{kid}, Learned Perceptual Image Patch Similarity (LPIPS)\cite{zhang2018unreasonableeffectivenessdeepfeatures}, Inception Score (IS)\cite{salimans2016improvedtechniquestraininggans}, Peak Signal-to-Noise Ratio (PSNR), and Structural Similarity Index (SSIM)\cite{ssim} to measure the overall performance numerically. The results in Table~\ref{Tab:quantitative} show that the proposed method outperforms other methods in most metrics.
\begin{table}[ht]
    \small
    \centering
        \begin{tabular}{ l|c|c|c|c|c|l }
            \hline
            \multicolumn{1}{c|}{Methods} &
            \multicolumn{1}{c|}{FID $\downarrow$} & 
            \multicolumn{1}{c|}{KID $\downarrow$} & 
            \multicolumn{1}{c|}{LPIPS $\downarrow$} &
            \multicolumn{1}{c|}{L1(\%) $\downarrow$} &
            \multicolumn{1}{c|}{PSNR $\uparrow$} & 
            \multicolumn{1}{c}{SSIM $\uparrow$} \\
            \hline
            \hline
            CycleGAN & 0.612 & 0.131 & 0.262 & 9.441 & 17.058 & 0.598 \\ %
            \hline
            PEIPNet & 1.222 & 0.087 & 0.149 & 3.965 & 24.584 & 0.763 \\
            \hline
            Pix2Pix & 1.106 & 0.088 & 0.147 & 3.193 & 23.940 & 0.823 \\ %
            \hline
            DeepFillv2 & \textbf{0.609} & 0.053 & 0.130 & 2.771 & 27.370 & 0.816 \\
            \hline
            SSDD-GAN (proposed) & 0.610 & \textbf{0.040} & \textbf{0.093}  & \textbf{2.296} & \textbf{28.896} & \textbf{0.878}\\
            \hline
        \end{tabular}
    \caption{\textbf{Quantitative Performance.} Comparison among various methods.}
    \label{Tab:quantitative}
\end{table}
\normalsize
Figure~\ref{fig:performance_overall_comparison} provides a detailed comparison of the aforementioned methods, evaluating their performance using L1(\%), PSNR, and SSIM metrics. These metrics collectively assess the accuracy, reconstruction quality, and structural consistency of each method. From Figure~\ref{fig:performance_overall_comparison}(a-c), we observe that our proposed method consistently outperforms competing models across all evaluated thresholds, demonstrating its robustness and superior reconstruction fidelity. Furthermore, Figure~\ref{fig:performance_overall_comparison}(d-f) highlights the effectiveness of our approach in handling varying mask sizes, showing that our method maintains higher accuracy and produces more reliable results even as the missing regions increase. These findings underscore the adaptability and generalization capability of our method compared to existing techniques.
Figure~\ref{fig:representative_samples} shows the randomly selected results of completing the missing region by our proposed method when compared with different models.
Figure~\ref{fig:ablation_study} shows the effects of sweeping across the number of diffusion steps $T$ on the L1(\%), PSNR, and SSIM metrics.
Finally, Figure~\ref{fig:synthesis} shows the surgical scene completion results of missing regions in the synthetic postmastoidectomy dataset via the zero-shot approach. For comparison, we selected the closest real surgical frames to evaluate the quality of the synthetic surgical scenes. By leveraging the precise camera pose information inherently generated within the synthetic postmastoidectomy dataset, our proposed method can fill the missing surgical scene that aligns well with the synthetic postmastoidectomy surface. This capability not only improves the realism of the synthetic surgical scenes but also represents a step forward in the surgical navigation field, with substantial potential to benefit a wide range of downstream tasks, including 3D scene understanding and anatomical structure tracking.
\section{Conclusion}
The proposed method effectively completes missing regions in complex and cluttered surgical scenes, addressing challenges such as irregular geometries and occlusions introduced by the random masking technique. The use of self-supervised learning makes our method highly adaptable and generalizable to other surgical domains. Furthermore, the fully synthetic postmastoidectomy scenes provide precise camera pose information for each synthetic microscopy surgical view, paving the way for future advancements in the field of image-guided cochlear implant surgery. One limitation of the proposed method is its suboptimal performance when dealing with large missing regions in an image. This limitation arises from the difficulty of restoring fine details and textures in large missing areas using small known regions, a challenge prevalent in surgical datasets with intricate anatomical structures and complex textures. Future work could explore methods to address this limitation, and leverage these synthetic complete surgical scene multi-views to develop methods for intraoperative navigation of anatomical structures and accurate surgical tool tracking, providing better surgical guidance and potentially improving surgical precision and outcomes. 
\midlacknowledgments{This work was supported in part by grants R01DC014037 and R01DC008408 from the NIDCD. This work is solely the responsibility of the authors and does not necessarily reflect the views of this institute.}

\bibliography{midl-samplebibliography}

\begin{thebibliography}{22}
\providecommand{\natexlab}[1]{#1}
\providecommand{\url}[1]{\texttt{#1}}
\expandafter\ifx\csname urlstyle\endcsname\relax
  \providecommand{\doi}[1]{doi: #1}\else
  \providecommand{\doi}{doi: \begingroup \urlstyle{rm}\Url}\fi

\bibitem[Bińkowski et~al.(2021)Bińkowski, Sutherland, Arbel, and Gretton]{kid}
Mikołaj Bińkowski, Danica~J. Sutherland, Michael Arbel, and Arthur Gretton.
\newblock Demystifying mmd gans, 2021.
\newblock URL \url{https://arxiv.org/abs/1801.01401}.

\bibitem[Daher et~al.(2023)Daher, Vasconcelos, and Stoyanov]{DAHER2023102994}
Rema Daher, Francisco Vasconcelos, and Danail Stoyanov.
\newblock A temporal learning approach to inpainting endoscopic specularities and its effect on image correspondence.
\newblock \emph{Medical Image Analysis}, 90:\penalty0 102994, 2023.
\newblock ISSN 1361-8415.
\newblock \doi{https://doi.org/10.1016/j.media.2023.102994}.
\newblock URL \url{https://www.sciencedirect.com/science/article/pii/S1361841523002542}.

\bibitem[Heusel et~al.(2018)Heusel, Ramsauer, Unterthiner, Nessler, and Hochreiter]{fid}
Martin Heusel, Hubert Ramsauer, Thomas Unterthiner, Bernhard Nessler, and Sepp Hochreiter.
\newblock Gans trained by a two time-scale update rule converge to a local nash equilibrium, 2018.
\newblock URL \url{https://arxiv.org/abs/1706.08500}.

\bibitem[Ho et~al.(2020)Ho, Jain, and Abbeel]{ho2020denoisingdiffusionprobabilisticmodels}
Jonathan Ho, Ajay Jain, and Pieter Abbeel.
\newblock Denoising diffusion probabilistic models, 2020.
\newblock URL \url{https://arxiv.org/abs/2006.11239}.

\bibitem[Iizuka et~al.(2017)Iizuka, Simo-Serra, and Ishikawa]{10.1145/3072959.3073659}
Satoshi Iizuka, Edgar Simo-Serra, and Hiroshi Ishikawa.
\newblock Globally and locally consistent image completion, July 2017.
\newblock ISSN 0730-0301.
\newblock URL \url{https://doi.org/10.1145/3072959.3073659}.

\bibitem[Isola et~al.(2018)Isola, Zhu, Zhou, and Efros]{isola2018imagetoimagetranslationconditionaladversarial}
Phillip Isola, Jun-Yan Zhu, Tinghui Zhou, and Alexei~A. Efros.
\newblock Image-to-image translation with conditional adversarial networks, 2018.
\newblock URL \url{https://arxiv.org/abs/1611.07004}.

\bibitem[Ko et~al.(2023)Ko, Choi, and Lee]{PEIPNet}
Jaekyun Ko, Wanuk Choi, and Sanghwan Lee.
\newblock Peipnet: Parametric efficient image-inpainting network with depthwise and pointwise convolution.
\newblock \emph{Sensors}, 23\penalty0 (19), 2023.
\newblock ISSN 1424-8220.
\newblock \doi{10.3390/s23198313}.
\newblock URL \url{https://www.mdpi.com/1424-8220/23/19/8313}.

\bibitem[Labadie and Noble(2018)]{labadie2018preliminary}
RF~Labadie and JH~Noble.
\newblock Preliminary results with image-guided cochlear implant insertion techniques.
\newblock \emph{Otol Neurotol}, 39\penalty0 (7):\penalty0 922--928, Aug 2018.
\newblock \doi{10.1097/MAO.0000000000001850}.

\bibitem[Park et~al.(2017)Park, Yang, Yumer, Ceylan, and Berg]{park2017transformationgroundedimagegenerationnetwork}
Eunbyung Park, Jimei Yang, Ersin Yumer, Duygu Ceylan, and Alexander~C. Berg.
\newblock Transformation-grounded image generation network for novel 3d view synthesis, 2017.
\newblock URL \url{https://arxiv.org/abs/1703.02921}.

\bibitem[Pathak et~al.(2016)Pathak, Krähenbühl, Donahue, Darrell, and Efros]{7780647}
Deepak Pathak, Philipp Krähenbühl, Jeff Donahue, Trevor Darrell, and Alexei~A. Efros.
\newblock Context encoders: Feature learning by inpainting.
\newblock In \emph{2016 IEEE Conference on Computer Vision and Pattern Recognition (CVPR)}, pages 2536--2544, 2016.
\newblock \doi{10.1109/CVPR.2016.278}.

\bibitem[Sabini and Rusak(2018)]{sabini2018paintingoutsideboximage}
Mark Sabini and Gili Rusak.
\newblock Painting outside the box: Image outpainting with gans, 2018.
\newblock URL \url{https://arxiv.org/abs/1808.08483}.

\bibitem[Salimans et~al.(2016)Salimans, Goodfellow, Zaremba, Cheung, Radford, and Chen]{salimans2016improvedtechniquestraininggans}
Tim Salimans, Ian Goodfellow, Wojciech Zaremba, Vicki Cheung, Alec Radford, and Xi~Chen.
\newblock Improved techniques for training gans, 2016.
\newblock URL \url{https://arxiv.org/abs/1606.03498}.

\bibitem[Suvorov et~al.(2021)Suvorov, Logacheva, Mashikhin, Remizova, Ashukha, Silvestrov, Kong, Goka, Park, and Lempitsky]{DBLP:journals/corr/abs-2109-07161}
Roman Suvorov, Elizaveta Logacheva, Anton Mashikhin, Anastasia Remizova, Arsenii Ashukha, Aleksei Silvestrov, Naejin Kong, Harshith Goka, Kiwoong Park, and Victor Lempitsky.
\newblock Resolution-robust large mask inpainting with fourier convolutions.
\newblock \emph{CoRR}, abs/2109.07161, 2021.
\newblock URL \url{https://arxiv.org/abs/2109.07161}.

\bibitem[Wang et~al.(2004)Wang, Bovik, Sheikh, and Simoncelli]{ssim}
Zhou Wang, A.C. Bovik, H.R. Sheikh, and E.P. Simoncelli.
\newblock Image quality assessment: from error visibility to structural similarity.
\newblock \emph{IEEE Transactions on Image Processing}, 13\penalty0 (4):\penalty0 600--612, 2004.
\newblock \doi{10.1109/TIP.2003.819861}.

\bibitem[Yu et~al.(2018)Yu, Lin, Yang, Shen, Lu, and Huang]{yu2018generativeimageinpaintingcontextual}
Jiahui Yu, Zhe Lin, Jimei Yang, Xiaohui Shen, Xin Lu, and Thomas~S. Huang.
\newblock Generative image inpainting with contextual attention, 2018.
\newblock URL \url{https://arxiv.org/abs/1801.07892}.

\bibitem[Yu et~al.(2019)Yu, Lin, Yang, Shen, Lu, and Huang]{yu2019freeformimageinpaintinggated}
Jiahui Yu, Zhe Lin, Jimei Yang, Xiaohui Shen, Xin Lu, and Thomas Huang.
\newblock Free-form image inpainting with gated convolution, 2019.
\newblock URL \url{https://arxiv.org/abs/1806.03589}.

\bibitem[Zhang et~al.(2018)Zhang, Isola, Efros, Shechtman, and Wang]{zhang2018unreasonableeffectivenessdeepfeatures}
Richard Zhang, Phillip Isola, Alexei~A. Efros, Eli Shechtman, and Oliver Wang.
\newblock The unreasonable effectiveness of deep features as a perceptual metric, 2018.
\newblock URL \url{https://arxiv.org/abs/1801.03924}.

\bibitem[Zhang and Noble(2024)]{zhang2024mastoidectomymultiviewsynthesissingle}
Yike Zhang and Jack Noble.
\newblock Mastoidectomy multi-view synthesis from a single microscopy image, 2024.
\newblock URL \url{https://arxiv.org/abs/2409.03190}.

\bibitem[Zhang and Noble(2023)]{10.1117/12.2655653}
Yike Zhang and Jack~H. Noble.
\newblock {Self-supervised registration and segmentation on ossicles with a single ground truth label}.
\newblock In Cristian~A. Linte and Jeffrey~H. Siewerdsen, editors, \emph{Medical Imaging 2023: Image-Guided Procedures, Robotic Interventions, and Modeling}, volume 12466, page 124660X. International Society for Optics and Photonics, SPIE, 2023.
\newblock \doi{10.1117/12.2655653}.
\newblock URL \url{https://doi.org/10.1117/12.2655653}.

\bibitem[Zhang et~al.(2024{\natexlab{a}})Zhang, Davalos, Su, Lou, and Noble]{10.1117/12.3008830}
Yike Zhang, Eduardo Davalos, Dingjie Su, Ange Lou, and Jack~H. Noble.
\newblock {Monocular microscope to CT registration using pose estimation of the incus for augmented reality cochlear implant surgery}.
\newblock In Jeffrey~H. Siewerdsen and Maryam~E. Rettmann, editors, \emph{Medical Imaging 2024: Image-Guided Procedures, Robotic Interventions, and Modeling}, volume 12928, page 129282I. International Society for Optics and Photonics, SPIE, 2024{\natexlab{a}}.
\newblock \doi{10.1117/12.3008830}.
\newblock URL \url{https://doi.org/10.1117/12.3008830}.

\bibitem[Zhang et~al.(2024{\natexlab{b}})Zhang, Davalos, Su, Lou, and Noble]{zhang2024mmunsupervisedmambabasedmastoidectomy}
Yike Zhang, Eduardo Davalos, Dingjie Su, Ange Lou, and Jack~H. Noble.
\newblock M\&m: Unsupervised mamba-based mastoidectomy for cochlear implant surgery with noisy data, 2024{\natexlab{b}}.
\newblock URL \url{https://arxiv.org/abs/2407.15787}.

\bibitem[Zhu et~al.(2020)Zhu, Park, Isola, and Efros]{zhu2020unpairedimagetoimagetranslationusing}
Jun-Yan Zhu, Taesung Park, Phillip Isola, and Alexei~A. Efros.
\newblock Unpaired image-to-image translation using cycle-consistent adversarial networks, 2020.
\newblock URL \url{https://arxiv.org/abs/1703.10593}.

\end{thebibliography}

\newpage
\appendix                                     
\section{Qualitative Results of SSDD-GAN}
\begin{figure}[ht]
  \centering
  \begin{minipage}{0.11\textwidth}
        \includegraphics[width=\textwidth]{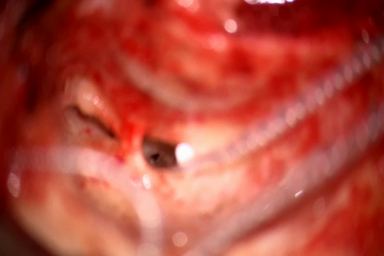}
    \end{minipage}
    \hfill
    \begin{minipage}{0.11\textwidth}
        \includegraphics[width=\textwidth]{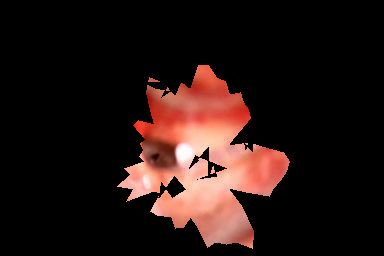}
    \end{minipage}
    \hfill
    \begin{minipage}{0.11\textwidth}
        \includegraphics[width=\textwidth]{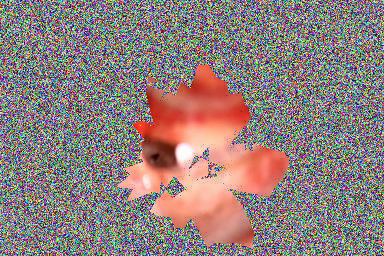}
    \end{minipage}
    \hfill
    \begin{minipage}{0.11\textwidth}
        \includegraphics[width=\textwidth]{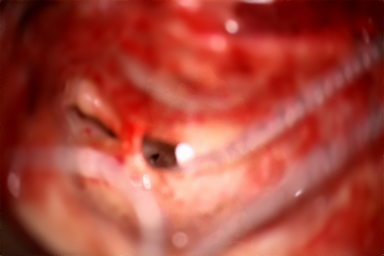}
    \end{minipage}
    \hspace{0.8em}
    \begin{minipage}{0.11\textwidth}
        \includegraphics[width=\textwidth]{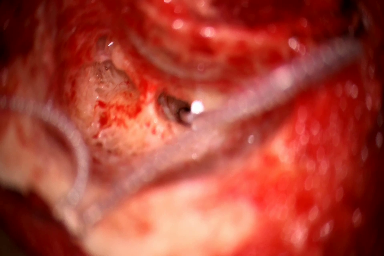}
    \end{minipage}
    \hfill
    \begin{minipage}{0.11\textwidth}
        \includegraphics[width=\textwidth]{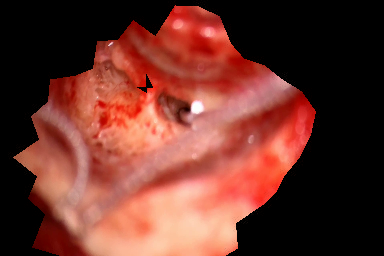}
    \end{minipage}
    \hfill
    \begin{minipage}{0.11\textwidth}
        \includegraphics[width=\textwidth]{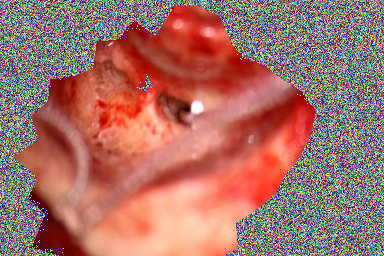}
    \end{minipage}
    \hfill
    \begin{minipage}{0.11\textwidth}
        \includegraphics[width=\textwidth]{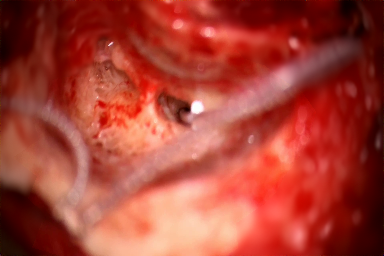}
    \end{minipage}
    \begin{minipage}{0.11\textwidth}
        \includegraphics[width=\textwidth]{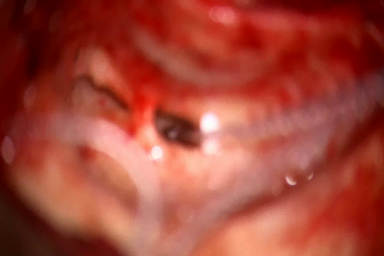}
    \end{minipage}
    \hfill
    \begin{minipage}{0.11\textwidth}
        \includegraphics[width=\textwidth]{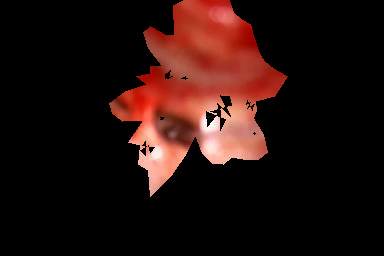}
    \end{minipage}
    \hfill
    \begin{minipage}{0.11\textwidth}
        \includegraphics[width=\textwidth]{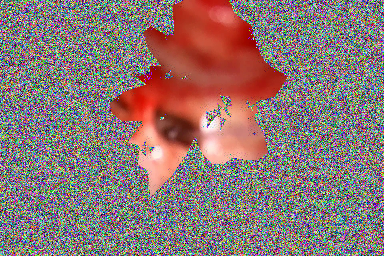}
    \end{minipage}
    \hfill
    \begin{minipage}{0.11\textwidth}
        \includegraphics[width=\textwidth]{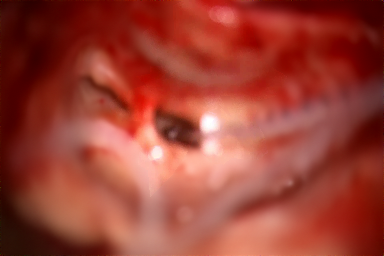}
    \end{minipage}
    \hspace{0.8em}
    \begin{minipage}{0.11\textwidth}
        \includegraphics[width=\textwidth]{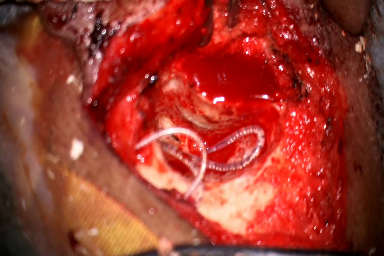}
    \end{minipage}
    \hfill
    \begin{minipage}{0.11\textwidth}
        \includegraphics[width=\textwidth]{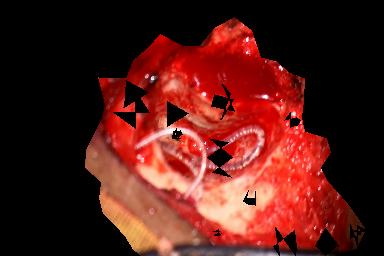}
    \end{minipage}
    \hfill
    \begin{minipage}{0.11\textwidth}
        \includegraphics[width=\textwidth]{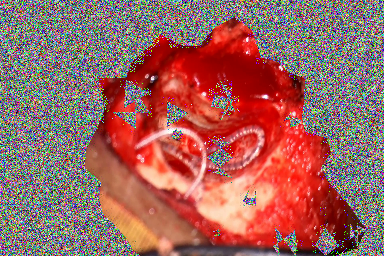}
    \end{minipage}
    \hfill
    \begin{minipage}{0.11\textwidth}
        \includegraphics[width=\textwidth]{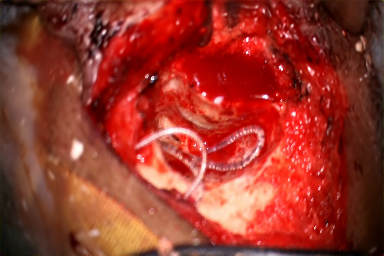}
    \end{minipage}
    \begin{minipage}{0.11\textwidth}
        \includegraphics[width=\textwidth]{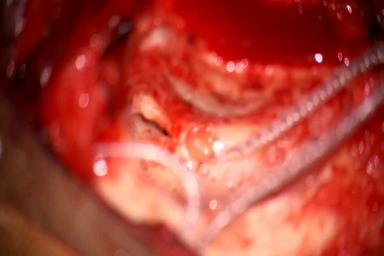}
    \end{minipage}
    \hfill
    \begin{minipage}{0.11\textwidth}
        \includegraphics[width=\textwidth]{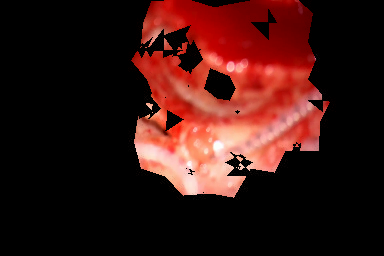}
    \end{minipage}
    \hfill
    \begin{minipage}{0.11\textwidth}
        \includegraphics[width=\textwidth]{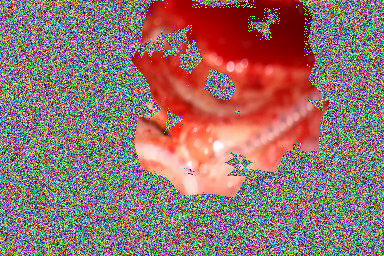}
    \end{minipage}
    \hfill
    \begin{minipage}{0.11\textwidth}
        \includegraphics[width=\textwidth]{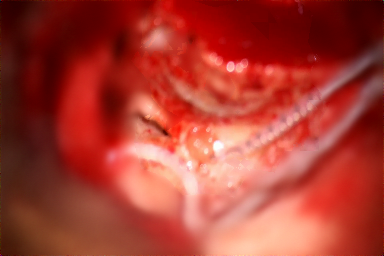}
    \end{minipage}
    \hspace{0.8em}
    \begin{minipage}{0.11\textwidth}
        \includegraphics[width=\textwidth]{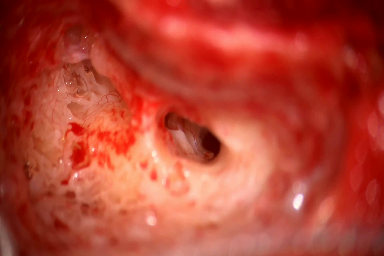}
    \end{minipage}
    \hfill
    \begin{minipage}{0.11\textwidth}
        \includegraphics[width=\textwidth]{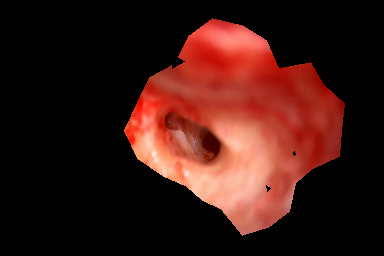}
    \end{minipage}
    \hfill
    \begin{minipage}{0.11\textwidth}
        \includegraphics[width=\textwidth]{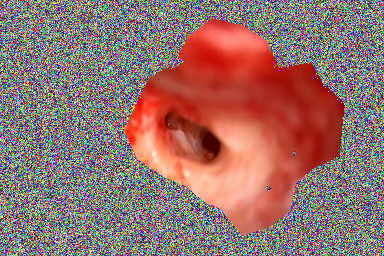}
    \end{minipage}
    \hfill
    \begin{minipage}{0.11\textwidth}
        \includegraphics[width=\textwidth]{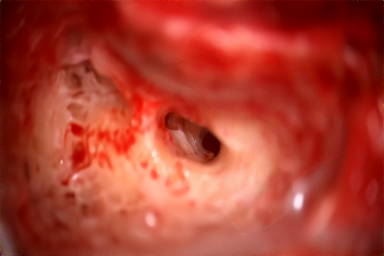}
    \end{minipage}
    \begin{minipage}{0.11\textwidth}
        \includegraphics[width=\textwidth]{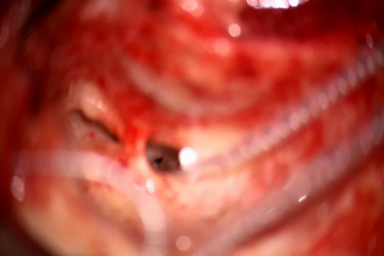}
    \end{minipage}
    \hfill
    \begin{minipage}{0.11\textwidth}
        \includegraphics[width=\textwidth]{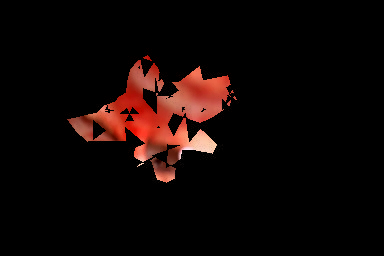}
    \end{minipage}
    \hfill
    \begin{minipage}{0.11\textwidth}
        \includegraphics[width=\textwidth]{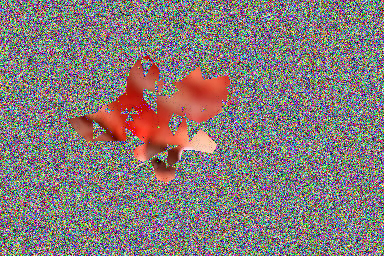}
    \end{minipage}
    \hfill
    \begin{minipage}{0.11\textwidth}
        \includegraphics[width=\textwidth]{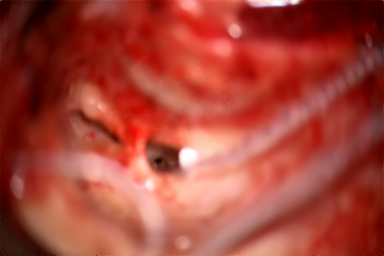}
    \end{minipage}
    \hspace{0.8em}
    \begin{minipage}{0.11\textwidth}
        \includegraphics[width=\textwidth]{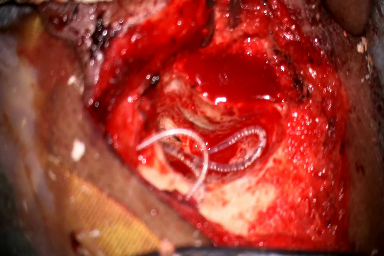}
    \end{minipage}
    \hfill
    \begin{minipage}{0.11\textwidth}
        \includegraphics[width=\textwidth]{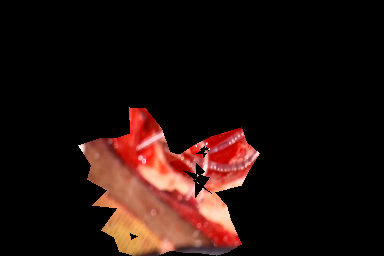}
    \end{minipage}
    \hfill
    \begin{minipage}{0.11\textwidth}
        \includegraphics[width=\textwidth]{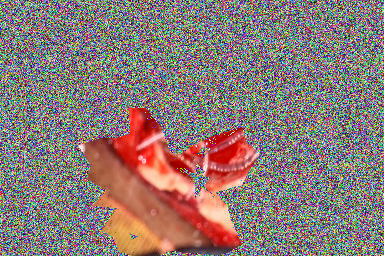}
    \end{minipage}
    \hfill
    \begin{minipage}{0.11\textwidth}
        \includegraphics[width=\textwidth]{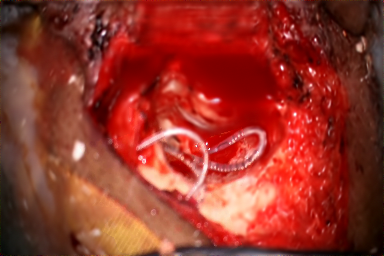}
    \end{minipage}
    \begin{minipage}{0.11\textwidth}
        \includegraphics[width=\textwidth]{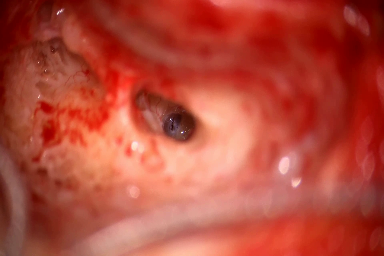}
    \end{minipage}
    \hfill
    \begin{minipage}{0.11\textwidth}
        \includegraphics[width=\textwidth]{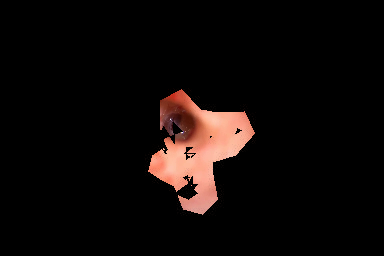}
    \end{minipage}
    \hfill
    \begin{minipage}{0.11\textwidth}
        \includegraphics[width=\textwidth]{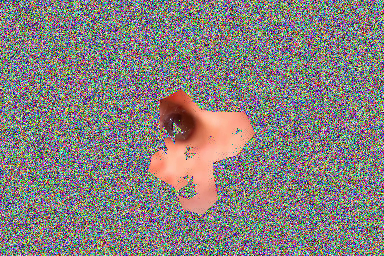}
    \end{minipage}
    \hfill
    \begin{minipage}{0.11\textwidth}
        \includegraphics[width=\textwidth]{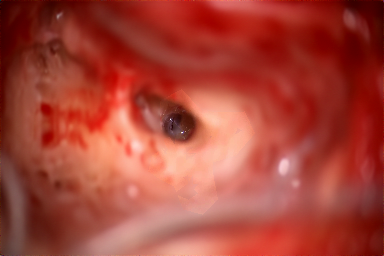}
    \end{minipage}
    \hspace{0.8em}
    \begin{minipage}{0.11\textwidth}
        \includegraphics[width=\textwidth]{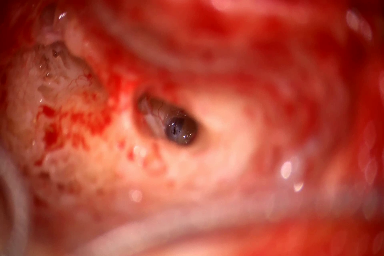}
    \end{minipage}
    \hfill
    \begin{minipage}{0.11\textwidth}
        \includegraphics[width=\textwidth]{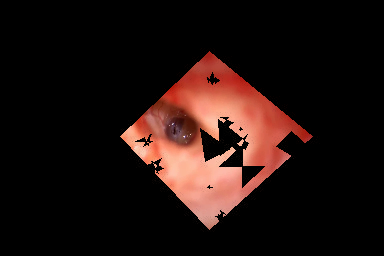}
    \end{minipage}
    \hfill
    \begin{minipage}{0.11\textwidth}
        \includegraphics[width=\textwidth]{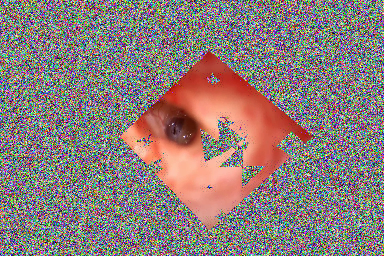}
    \end{minipage}
    \hfill
    \begin{minipage}{0.11\textwidth}
        \includegraphics[width=\textwidth]{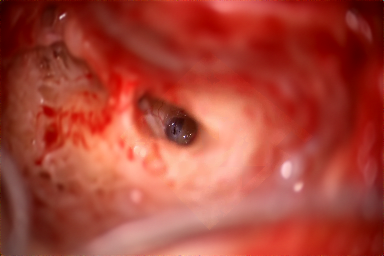}
    \end{minipage}
    \begin{minipage}{0.11\textwidth}
        \includegraphics[width=\textwidth]{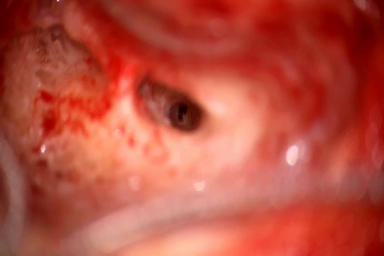}
    \end{minipage}
    \hfill
    \begin{minipage}{0.11\textwidth}
        \includegraphics[width=\textwidth]{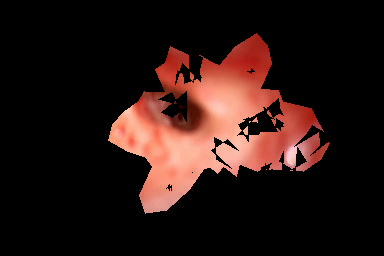}
    \end{minipage}
    \hfill
    \begin{minipage}{0.11\textwidth}
        \includegraphics[width=\textwidth]{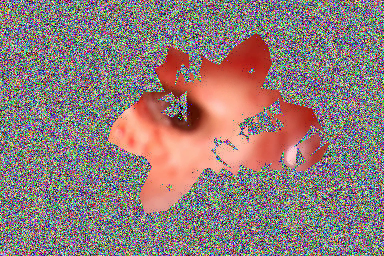}
    \end{minipage}
    \hfill
    \begin{minipage}{0.11\textwidth}
        \includegraphics[width=\textwidth]{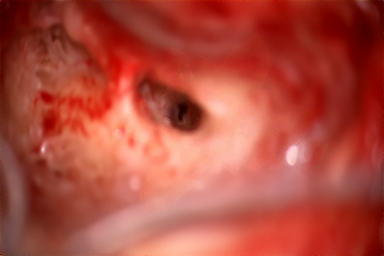}
    \end{minipage}
    \hspace{0.8em}
    \begin{minipage}{0.11\textwidth}
        \includegraphics[width=\textwidth]{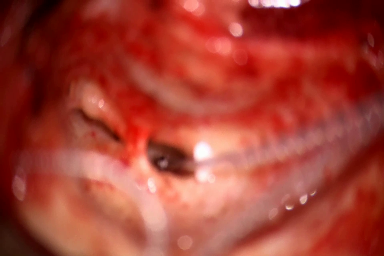}
    \end{minipage}
    \hfill
    \begin{minipage}{0.11\textwidth}
        \includegraphics[width=\textwidth]{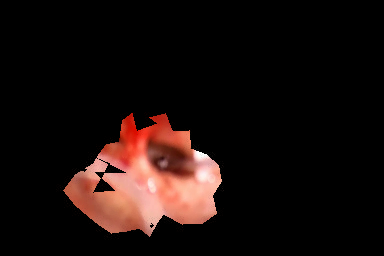}
    \end{minipage}
    \hfill
    \begin{minipage}{0.11\textwidth}
        \includegraphics[width=\textwidth]{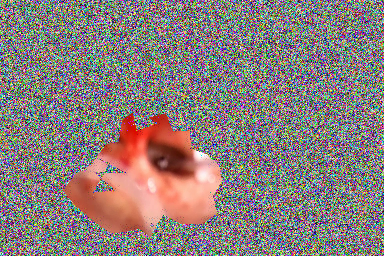}
    \end{minipage}
    \hfill
    \begin{minipage}{0.11\textwidth}
        \includegraphics[width=\textwidth]{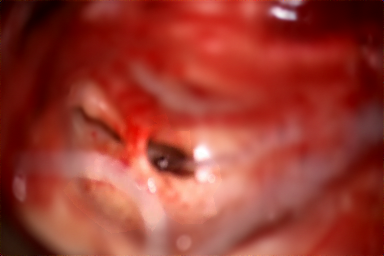}
    \end{minipage}
    \begin{minipage}{0.11\textwidth}
        \includegraphics[width=\textwidth]{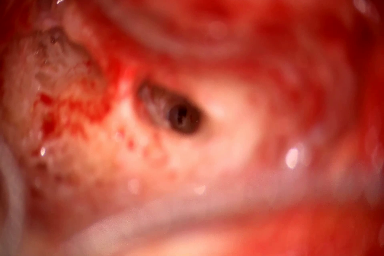}
    \end{minipage}
    \hfill
    \begin{minipage}{0.11\textwidth}
        \includegraphics[width=\textwidth]{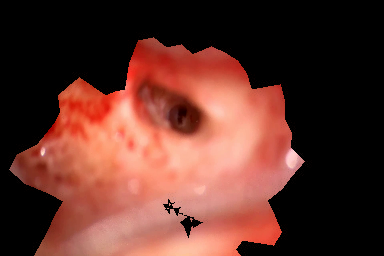}
    \end{minipage}
    \hfill
    \begin{minipage}{0.11\textwidth}
        \includegraphics[width=\textwidth]{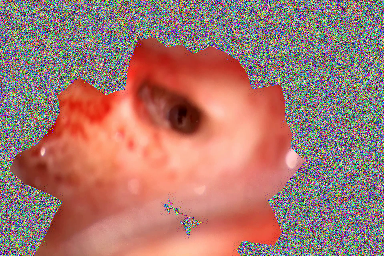}
    \end{minipage}
    \hfill
    \begin{minipage}{0.11\textwidth}
        \includegraphics[width=\textwidth]{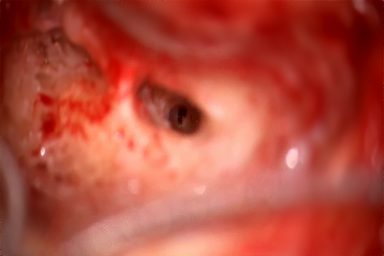}
    \end{minipage}
    \hspace{0.8em}
    \begin{minipage}{0.11\textwidth}
        \includegraphics[width=\textwidth]{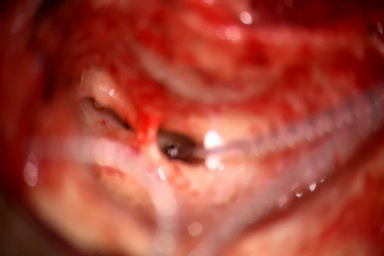}
    \end{minipage}
    \hfill
    \begin{minipage}{0.11\textwidth}
        \includegraphics[width=\textwidth]{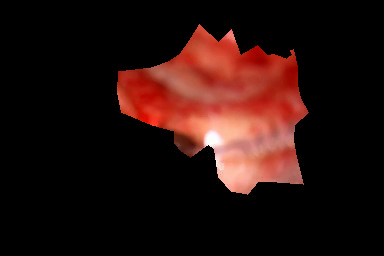}
    \end{minipage}
    \hfill
    \begin{minipage}{0.11\textwidth}
        \includegraphics[width=\textwidth]{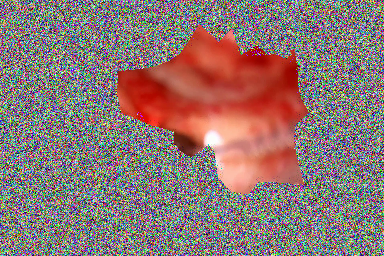}
    \end{minipage}
    \hfill
    \begin{minipage}{0.11\textwidth}
        \includegraphics[width=\textwidth]{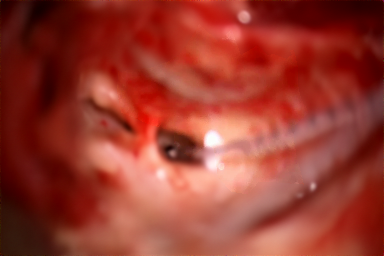}
    \end{minipage}
    \begin{minipage}{0.11\textwidth}
        \includegraphics[width=\textwidth]{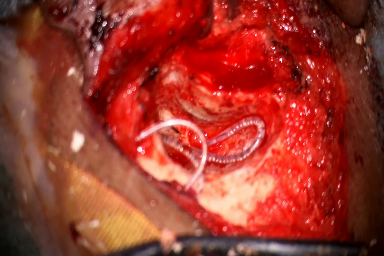}
    \end{minipage}
    \hfill
    \begin{minipage}{0.11\textwidth}
        \includegraphics[width=\textwidth]{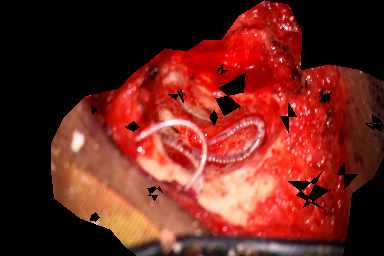}
    \end{minipage}
    \hfill
    \begin{minipage}{0.11\textwidth}
        \includegraphics[width=\textwidth]{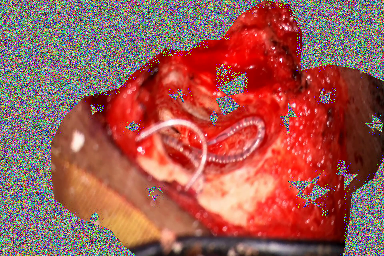}
    \end{minipage}
    \hfill
    \begin{minipage}{0.11\textwidth}
        \includegraphics[width=\textwidth]{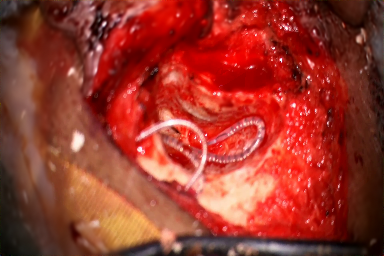}
    \end{minipage}
    \hspace{0.8em}
    \begin{minipage}{0.11\textwidth}
        \includegraphics[width=\textwidth]{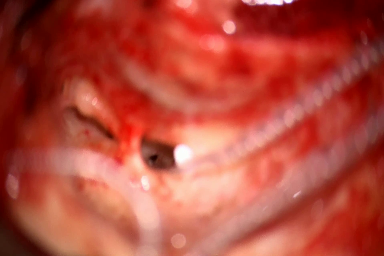}
    \end{minipage}
    \hfill
    \begin{minipage}{0.11\textwidth}
        \includegraphics[width=\textwidth]{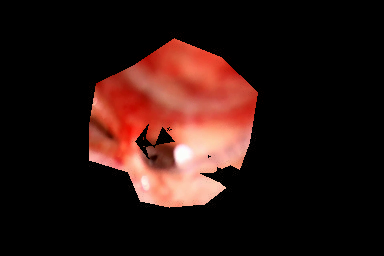}
    \end{minipage}
    \hfill
    \begin{minipage}{0.11\textwidth}
        \includegraphics[width=\textwidth]{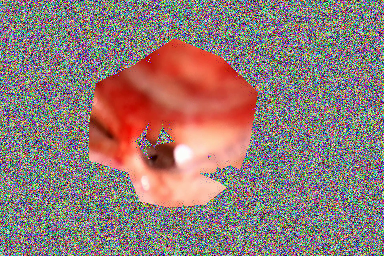}
    \end{minipage}
    \hfill
    \begin{minipage}{0.11\textwidth}
        \includegraphics[width=\textwidth]{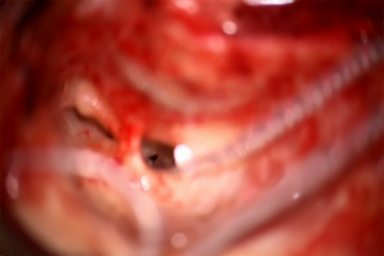}
    \end{minipage}
    \begin{minipage}{0.11\textwidth}
        \includegraphics[width=\textwidth]{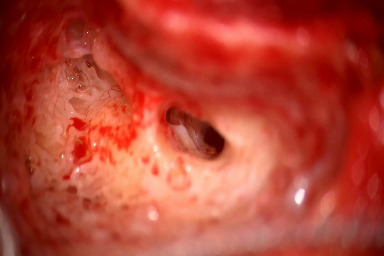}
    \end{minipage}
    \hfill
    \begin{minipage}{0.11\textwidth}
        \includegraphics[width=\textwidth]{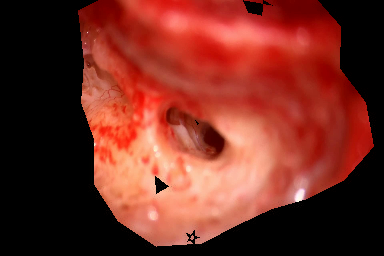}
    \end{minipage}
    \hfill
    \begin{minipage}{0.11\textwidth}
        \includegraphics[width=\textwidth]{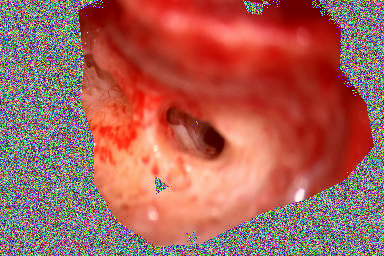}
    \end{minipage}
    \hfill
    \begin{minipage}{0.11\textwidth}
        \includegraphics[width=\textwidth]{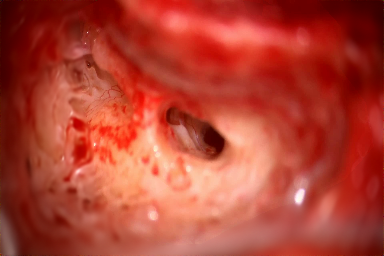}
    \end{minipage}
    \hspace{0.8em}
    \begin{minipage}{0.11\textwidth}
        \includegraphics[width=\textwidth]{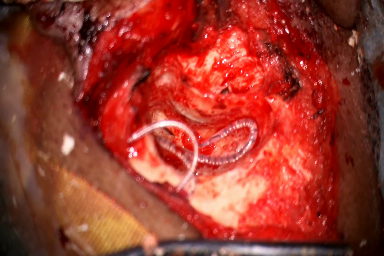}
    \end{minipage}
    \hfill
    \begin{minipage}{0.11\textwidth}
        \includegraphics[width=\textwidth]{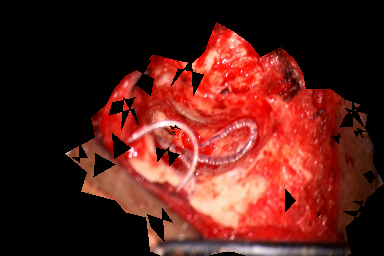}
    \end{minipage}
    \hfill
    \begin{minipage}{0.11\textwidth}
        \includegraphics[width=\textwidth]{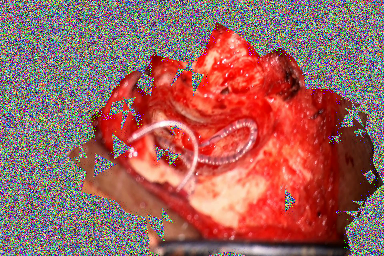}
    \end{minipage}
    \hfill
    \begin{minipage}{0.11\textwidth}
        \includegraphics[width=\textwidth]{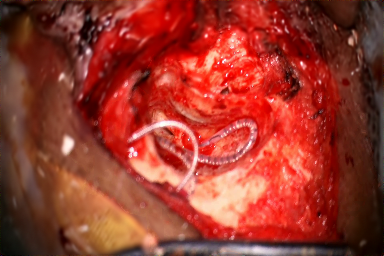}
    \end{minipage}
    \begin{minipage}{0.11\textwidth}
        \includegraphics[width=\textwidth]{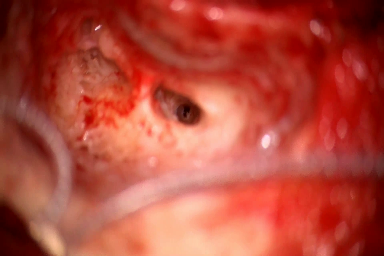}
    \end{minipage}
    \hfill
    \begin{minipage}{0.11\textwidth}
        \includegraphics[width=\textwidth]{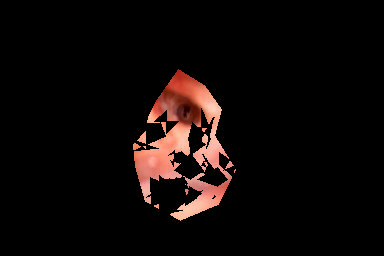}
    \end{minipage}
    \hfill
    \begin{minipage}{0.11\textwidth}
        \includegraphics[width=\textwidth]{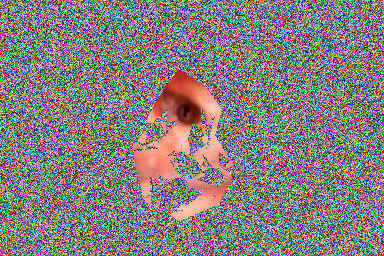}
    \end{minipage}
    \hfill
    \begin{minipage}{0.11\textwidth}
        \includegraphics[width=\textwidth]{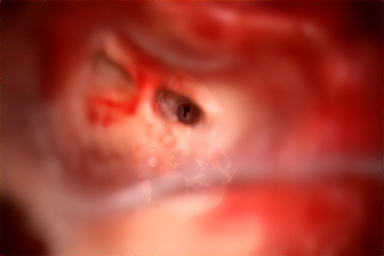}
    \end{minipage}
    \hspace{0.8em}
    \begin{minipage}{0.11\textwidth}
        \includegraphics[width=\textwidth]{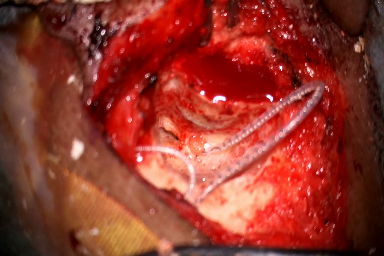}
    \end{minipage}
    \hfill
    \begin{minipage}{0.11\textwidth}
        \includegraphics[width=\textwidth]{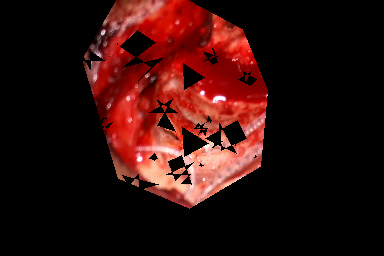}
    \end{minipage}
    \hfill
    \begin{minipage}{0.11\textwidth}
        \includegraphics[width=\textwidth]{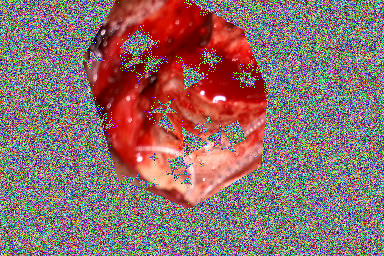}
    \end{minipage}
    \hfill
    \begin{minipage}{0.11\textwidth}
        \includegraphics[width=\textwidth]{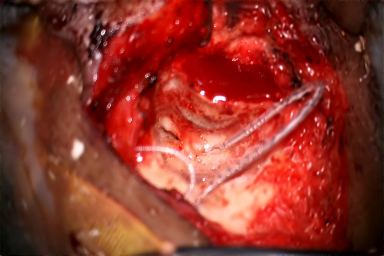}
    \end{minipage}
    \begin{minipage}{0.11\textwidth}
        \includegraphics[width=\textwidth]{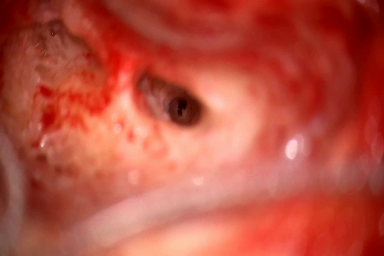}
    \end{minipage}
    \hfill
    \begin{minipage}{0.11\textwidth}
        \includegraphics[width=\textwidth]{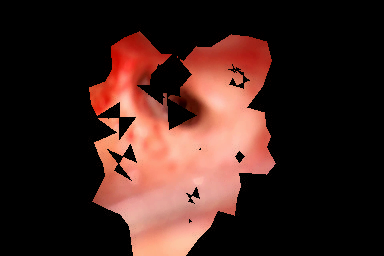}
    \end{minipage}
    \hfill
    \begin{minipage}{0.11\textwidth}
        \includegraphics[width=\textwidth]{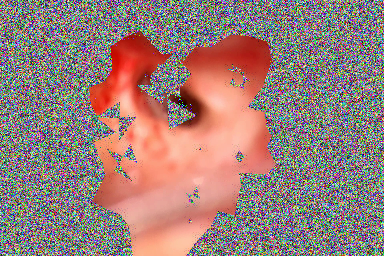}
    \end{minipage}
    \hfill
    \begin{minipage}{0.11\textwidth}
        \includegraphics[width=\textwidth]{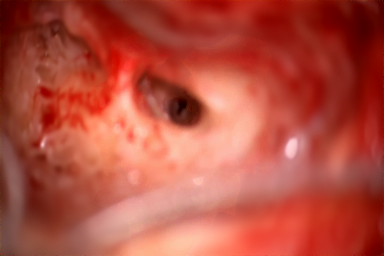}
    \end{minipage}
    \hspace{0.8em}
    \begin{minipage}{0.11\textwidth}
        \includegraphics[width=\textwidth]{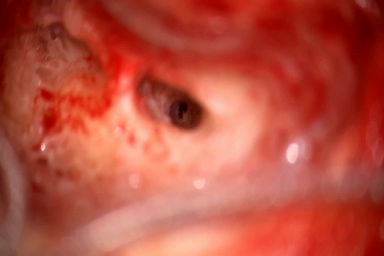}
    \end{minipage}
    \hfill
    \begin{minipage}{0.11\textwidth}
        \includegraphics[width=\textwidth]{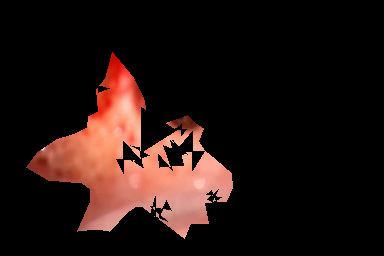}
    \end{minipage}
    \hfill
    \begin{minipage}{0.11\textwidth}
        \includegraphics[width=\textwidth]{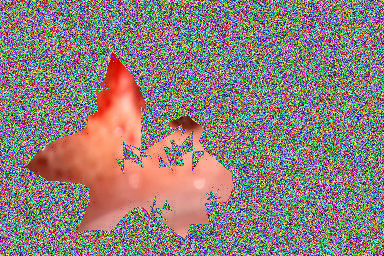}
    \end{minipage}
    \hfill
    \begin{minipage}{0.11\textwidth}
        \includegraphics[width=\textwidth]{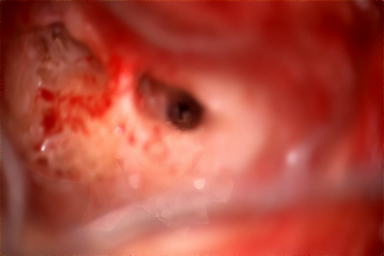}
    \end{minipage}
    \begin{minipage}{0.11\textwidth}
        \includegraphics[width=\textwidth]{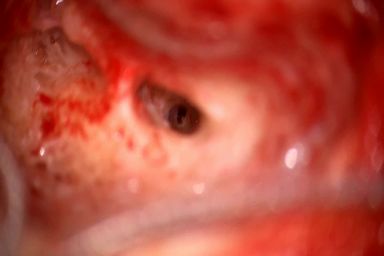}
    \end{minipage}
    \hfill
    \begin{minipage}{0.11\textwidth}
        \includegraphics[width=\textwidth]{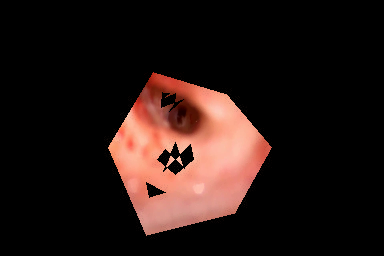}
    \end{minipage}
    \hfill
    \begin{minipage}{0.11\textwidth}
        \includegraphics[width=\textwidth]{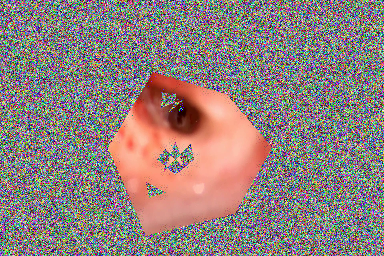}
    \end{minipage}
    \hfill
    \begin{minipage}{0.11\textwidth}
        \includegraphics[width=\textwidth]{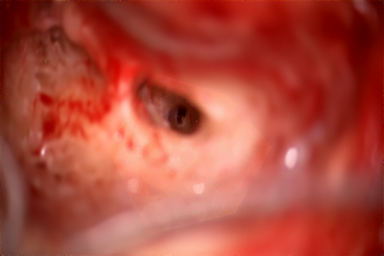}
    \end{minipage}
    \hspace{0.8em}
    \begin{minipage}{0.11\textwidth}
        \includegraphics[width=\textwidth]{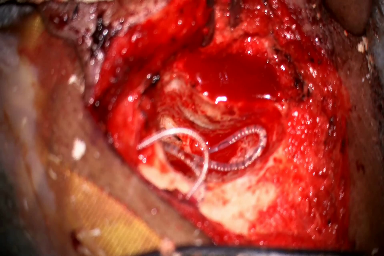}
    \end{minipage}
    \hfill
    \begin{minipage}{0.11\textwidth}
        \includegraphics[width=\textwidth]{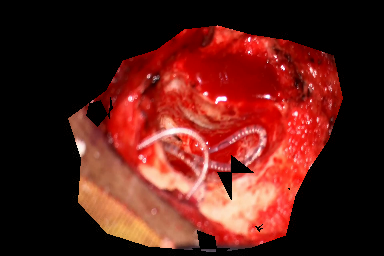}
    \end{minipage}
    \hfill
    \begin{minipage}{0.11\textwidth}
        \includegraphics[width=\textwidth]{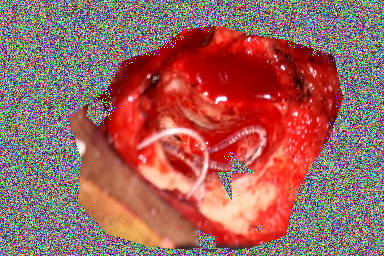}
    \end{minipage}
    \hfill
    \begin{minipage}{0.11\textwidth}
        \includegraphics[width=\textwidth]{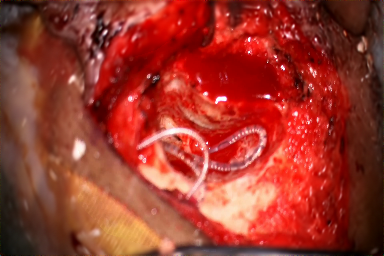}
    \end{minipage}
    \begin{minipage}{0.11\textwidth}
        \includegraphics[width=\textwidth]{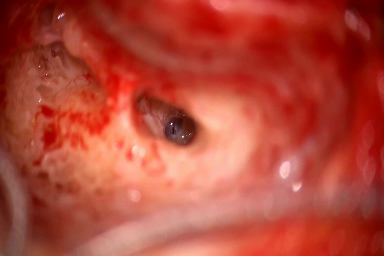}
    \end{minipage}
    \hfill
    \begin{minipage}{0.11\textwidth}
        \includegraphics[width=\textwidth]{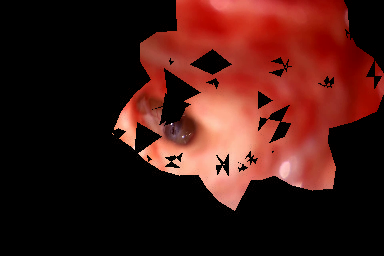}
    \end{minipage}
    \hfill
    \begin{minipage}{0.11\textwidth}
        \includegraphics[width=\textwidth]{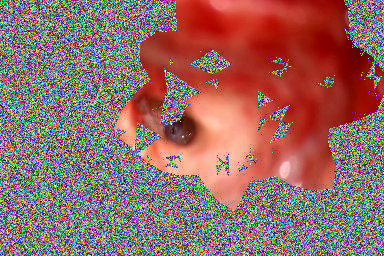}
    \end{minipage}
    \hfill
    \begin{minipage}{0.11\textwidth}
        \includegraphics[width=\textwidth]{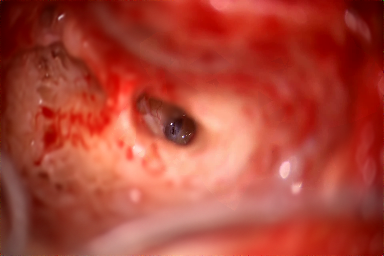}
    \end{minipage}
    \hspace{0.8em}
    \begin{minipage}{0.11\textwidth}
        \includegraphics[width=\textwidth]{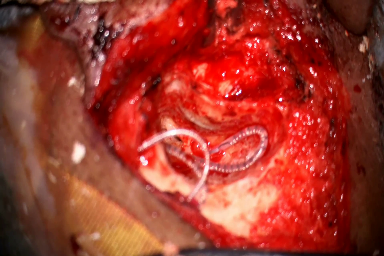}
    \end{minipage}
    \hfill
    \begin{minipage}{0.11\textwidth}
        \includegraphics[width=\textwidth]{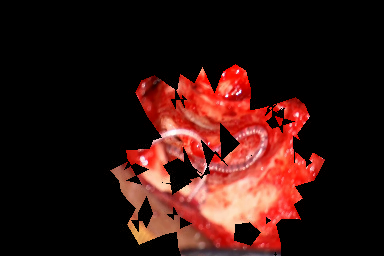}
    \end{minipage}
    \hfill
    \begin{minipage}{0.11\textwidth}
        \includegraphics[width=\textwidth]{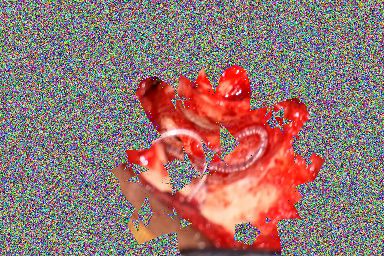}
    \end{minipage}
    \hfill
    \begin{minipage}{0.11\textwidth}
        \includegraphics[width=\textwidth]{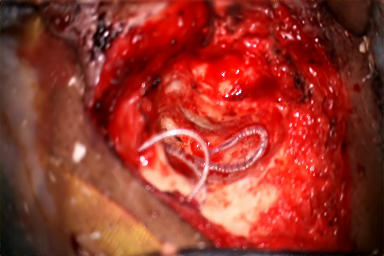}
    \end{minipage}
    \begin{minipage}{0.11\textwidth}
        \includegraphics[width=\textwidth]{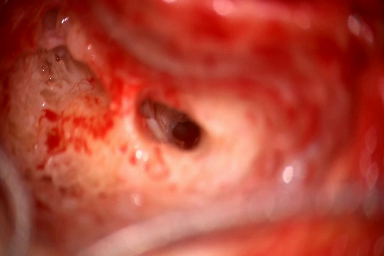}
    \end{minipage}
    \hfill
    \begin{minipage}{0.11\textwidth}
        \includegraphics[width=\textwidth]{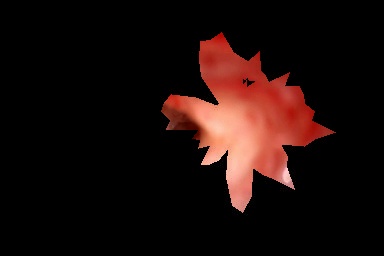}
    \end{minipage}
    \hfill
    \begin{minipage}{0.11\textwidth}
        \includegraphics[width=\textwidth]{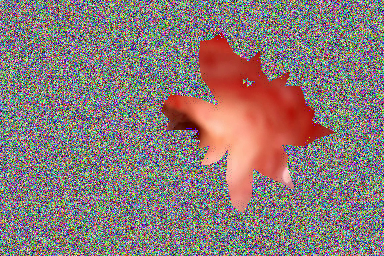}
    \end{minipage}
    \hfill
    \begin{minipage}{0.11\textwidth}
        \includegraphics[width=\textwidth]{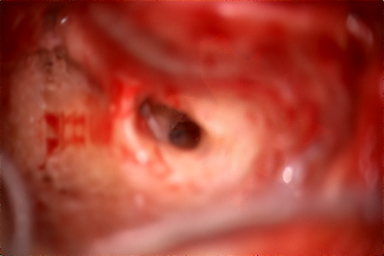}
    \end{minipage}
    \hspace{0.8em}
    \begin{minipage}{0.11\textwidth}
        \includegraphics[width=\textwidth]{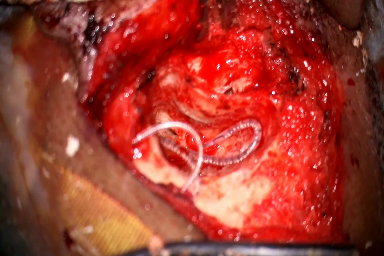}
    \end{minipage}
    \hfill
    \begin{minipage}{0.11\textwidth}
        \includegraphics[width=\textwidth]{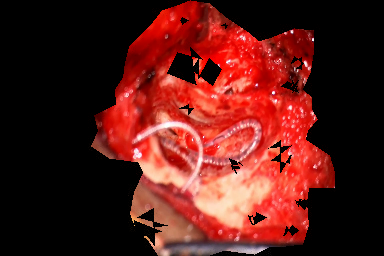}
    \end{minipage}
    \hfill
    \begin{minipage}{0.11\textwidth}
        \includegraphics[width=\textwidth]{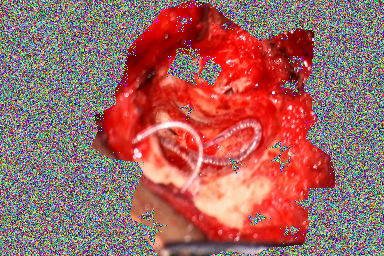}
    \end{minipage}
    \hfill
    \begin{minipage}{0.11\textwidth}
        \includegraphics[width=\textwidth]{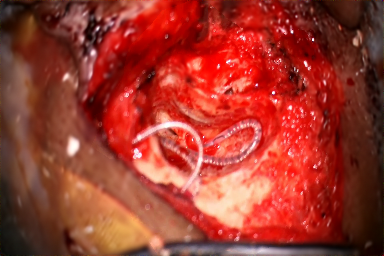}
    \end{minipage}
    \begin{minipage}{0.11\textwidth}
        \centering
        \includegraphics[width=\textwidth]{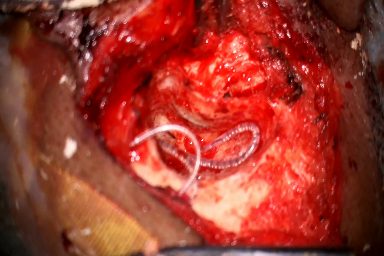}
        \footnotesize{(a)}
    \end{minipage}
    \hfill
    \begin{minipage}{0.11\textwidth}
        \centering
        \includegraphics[width=\textwidth]{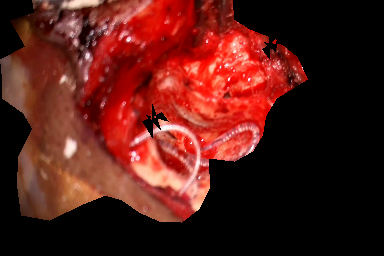}
        \footnotesize{(b)}
    \end{minipage}
    \hfill
    \begin{minipage}{0.11\textwidth}
        \centering
        \includegraphics[width=\textwidth]{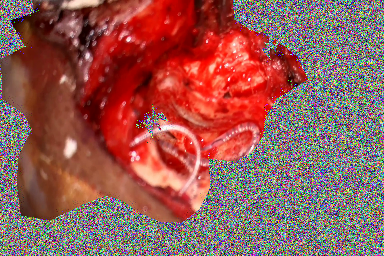}
        \footnotesize{(c)}
    \end{minipage}
    \hfill
    \begin{minipage}{0.11\textwidth}
        \centering
        \includegraphics[width=\textwidth]{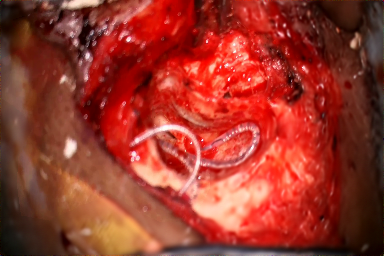}
        \footnotesize{(d)}
    \end{minipage}
    \hspace{0.8em}
    \begin{minipage}{0.11\textwidth}
        \centering
        \includegraphics[width=\textwidth]{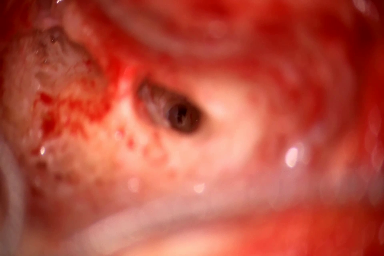}
        \footnotesize{(a)}
    \end{minipage}
    \hfill
    \begin{minipage}{0.11\textwidth}
        \centering
        \includegraphics[width=\textwidth]{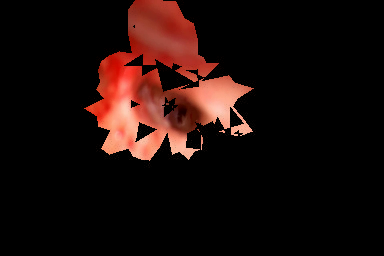}
        \footnotesize{(b)}
    \end{minipage}
    \hfill
    \begin{minipage}{0.11\textwidth}
        \centering
        \includegraphics[width=\textwidth]{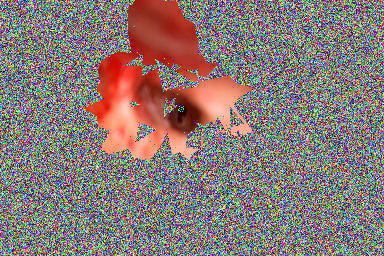}
        \footnotesize{(c)}
    \end{minipage}
    \hfill
    \begin{minipage}{0.11\textwidth}
        \centering
        \includegraphics[width=\textwidth]{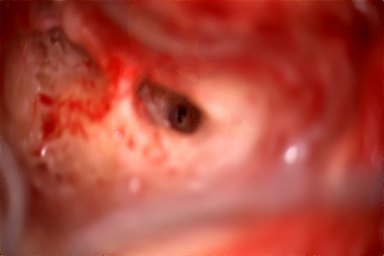}
        \footnotesize{(d)}
    \end{minipage}
\caption{\textbf{Qualitative Performance Evaluation}. (a) Original Image. (b) Masked Image. (c) Diffused Image. (d) Reconstructed Image.}
\label{fig:qualitative}
\end{figure}

Figure~\ref{fig:qualitative} demonstrates the effectiveness of the proposed framework in restoring various missing surgical scenes across different mask ratios.

\newpage
\section{More Qualitative Comparisons}
\begin{figure*}[ht]
    \begin{minipage}[t]{0.24\textwidth}
        \begin{minipage}{0.50\textwidth}
            \includegraphics[width=\textwidth]{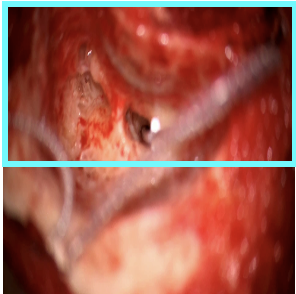}
        \end{minipage}\hspace{-0.3em}
        \begin{minipage}{0.50\textwidth}
            \includegraphics[width=\textwidth]{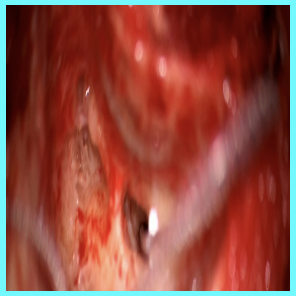}
        \end{minipage}
    \end{minipage}
    \hfill
    \begin{minipage}[t]{0.24\textwidth}
        \begin{minipage}{0.50\textwidth}
            \includegraphics[width=\textwidth]{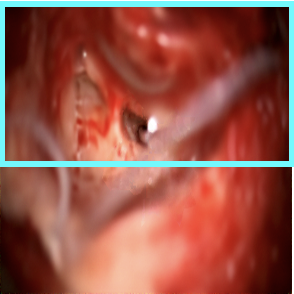}
        \end{minipage}\hspace{-0.3em}
        \begin{minipage}{0.50\textwidth}
            \includegraphics[width=\textwidth]{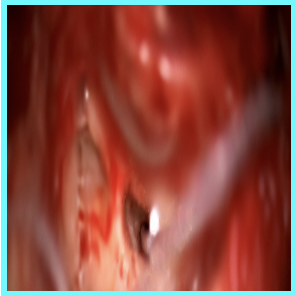}
        \end{minipage}
    \end{minipage}
    \hfill
    \begin{minipage}[t]{0.24\textwidth}
        \begin{minipage}{0.50\textwidth}
            \includegraphics[width=\textwidth]{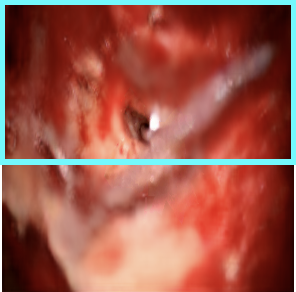}
        \end{minipage}\hspace{-0.3em}
        \begin{minipage}{0.50\textwidth}
            \includegraphics[width=\textwidth]{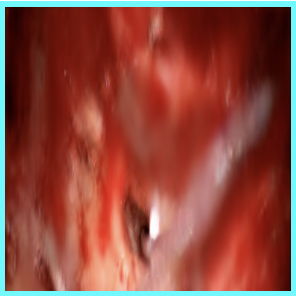}
        \end{minipage}
    \end{minipage}
    \hfill
    \begin{minipage}[t]{0.24\textwidth}
        \begin{minipage}{0.50\textwidth}
            \includegraphics[width=\textwidth]{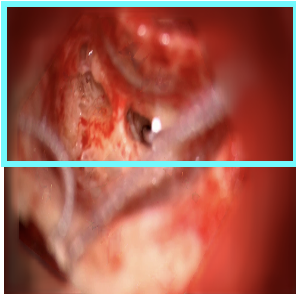}
        \end{minipage}\hspace{-0.3em}
        \begin{minipage}{0.50\textwidth}
            \includegraphics[width=\textwidth]{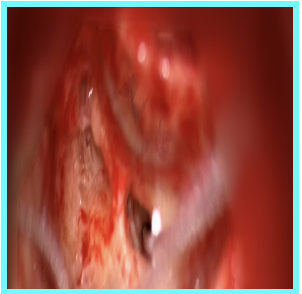}
        \end{minipage}
    \end{minipage}
    \hfill

    \begin{minipage}[t]{0.24\textwidth}
        \begin{minipage}{0.50\textwidth}
            \includegraphics[width=\textwidth]{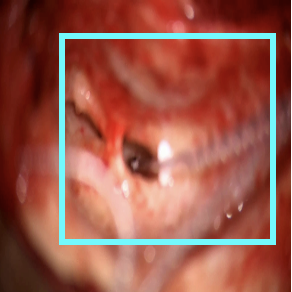}
        \end{minipage}\hspace{-0.3em}
        \begin{minipage}{0.50\textwidth}
            \includegraphics[width=\textwidth]{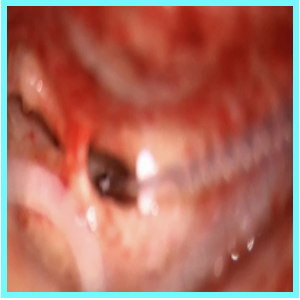}
        \end{minipage}
    \end{minipage}
    \hfill
    \begin{minipage}[t]{0.24\textwidth}
        \begin{minipage}{0.50\textwidth}
            \includegraphics[width=\textwidth]{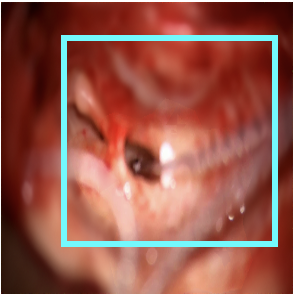}
        \end{minipage}\hspace{-0.3em}
        \begin{minipage}{0.50\textwidth}
            \includegraphics[width=\textwidth]{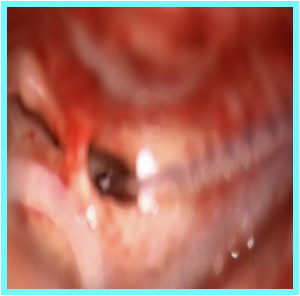}
        \end{minipage}
    \end{minipage}
    \hfill
    \begin{minipage}[t]{0.24\textwidth}
        \begin{minipage}{0.50\textwidth}
            \includegraphics[width=\textwidth]{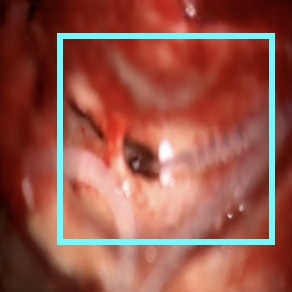}
        \end{minipage}\hspace{-0.3em}
        \begin{minipage}{0.50\textwidth}
            \includegraphics[width=\textwidth]{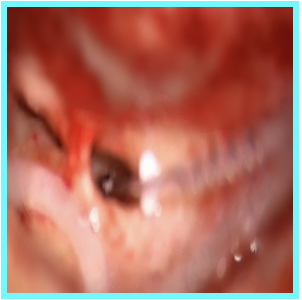}
        \end{minipage}
    \end{minipage}
    \hfill
    \begin{minipage}[t]{0.24\textwidth}
        \begin{minipage}{0.50\textwidth}
            \includegraphics[width=\textwidth]{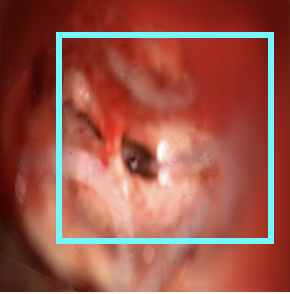}
        \end{minipage}\hspace{-0.3em}
        \begin{minipage}{0.50\textwidth}
            \includegraphics[width=\textwidth]{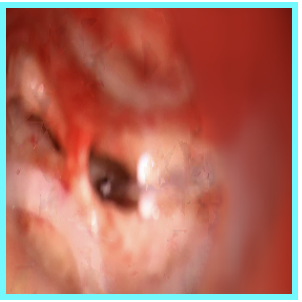}
        \end{minipage}
    \end{minipage}
    \hfill

    \begin{minipage}[t]{0.24\textwidth}
        \begin{minipage}{0.50\textwidth}
            \includegraphics[width=\textwidth]{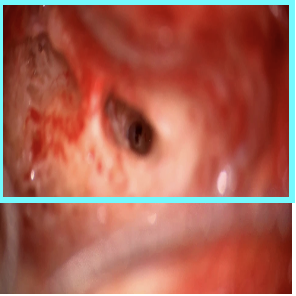}
        \end{minipage}\hspace{-0.3em}
        \begin{minipage}{0.50\textwidth}
            \includegraphics[width=\textwidth]{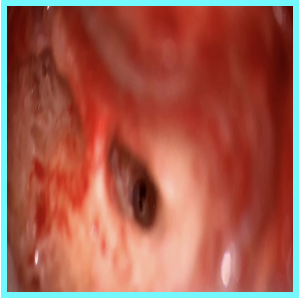}
        \end{minipage}
    \end{minipage}
    \hfill
    \begin{minipage}[t]{0.24\textwidth}
        \begin{minipage}{0.50\textwidth}
            \includegraphics[width=\textwidth]{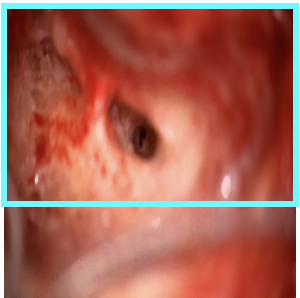}
        \end{minipage}\hspace{-0.3em}
        \begin{minipage}{0.50\textwidth}
            \includegraphics[width=\textwidth]{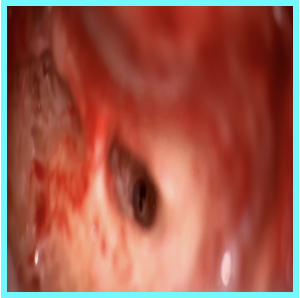}
        \end{minipage}
    \end{minipage}
    \hfill
    \begin{minipage}[t]{0.24\textwidth}
        \begin{minipage}{0.50\textwidth}
            \includegraphics[width=\textwidth]{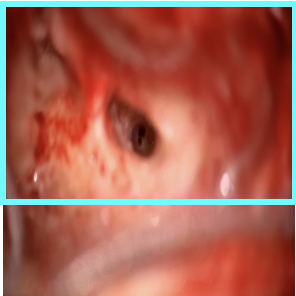}
        \end{minipage}\hspace{-0.3em}
        \begin{minipage}{0.50\textwidth}
            \includegraphics[width=\textwidth]{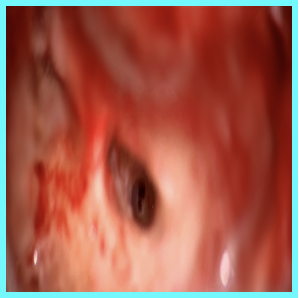}
        \end{minipage}
    \end{minipage}
    \hfill
    \begin{minipage}[t]{0.24\textwidth}
        \begin{minipage}{0.50\textwidth}
            \includegraphics[width=\textwidth]{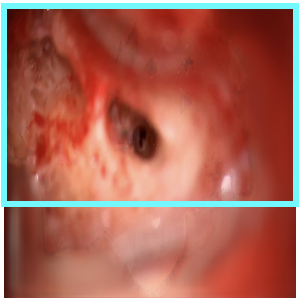}
        \end{minipage}\hspace{-0.3em}
        \begin{minipage}{0.50\textwidth}
            \includegraphics[width=\textwidth]{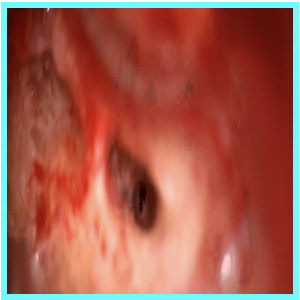}
        \end{minipage}
    \end{minipage}
    \hfill

    \begin{minipage}[t]{0.24\textwidth}
        \begin{minipage}{0.50\textwidth}
            \includegraphics[width=\textwidth]{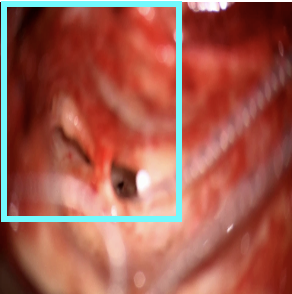}
        \end{minipage}\hspace{-0.3em}
        \begin{minipage}{0.50\textwidth}
            \includegraphics[width=\textwidth]{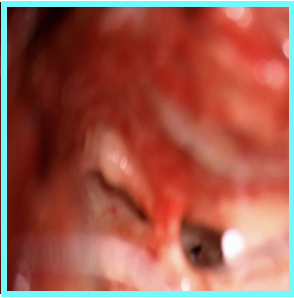}
        \end{minipage}
    \end{minipage}
    \hfill
    \begin{minipage}[t]{0.24\textwidth}
        \begin{minipage}{0.50\textwidth}
            \includegraphics[width=\textwidth]{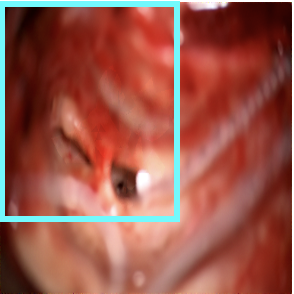}
        \end{minipage}\hspace{-0.3em}
        \begin{minipage}{0.50\textwidth}
            \includegraphics[width=\textwidth]{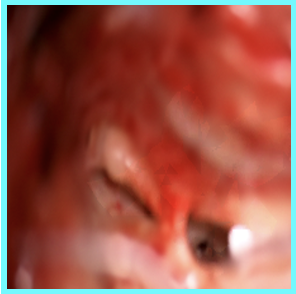}
        \end{minipage}
    \end{minipage}
    \hfill
    \begin{minipage}[t]{0.24\textwidth}
        \begin{minipage}{0.50\textwidth}
            \includegraphics[width=\textwidth]{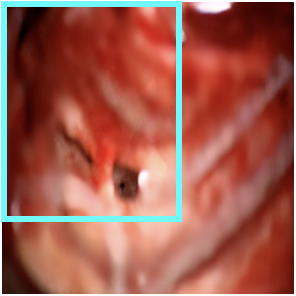}
        \end{minipage}\hspace{-0.3em}
        \begin{minipage}{0.50\textwidth}
            \includegraphics[width=\textwidth]{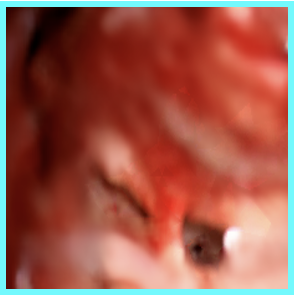}
        \end{minipage}
    \end{minipage}
    \hfill
    \begin{minipage}[t]{0.24\textwidth}
        \begin{minipage}{0.50\textwidth}
            \includegraphics[width=\textwidth]{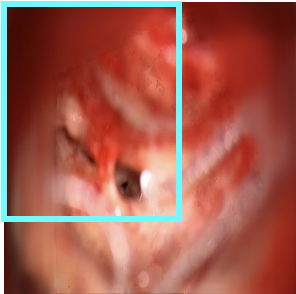}
        \end{minipage}\hspace{-0.3em}
        \begin{minipage}{0.50\textwidth}
            \includegraphics[width=\textwidth]{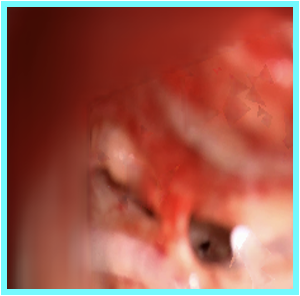}
        \end{minipage}
    \end{minipage}
    \hfill

    \begin{minipage}[t]{0.24\textwidth}
        \begin{minipage}{0.50\textwidth}
            \includegraphics[width=\textwidth]{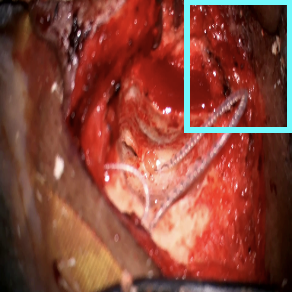}
        \end{minipage}\hspace{-0.3em}
        \begin{minipage}{0.50\textwidth}
            \includegraphics[width=\textwidth]{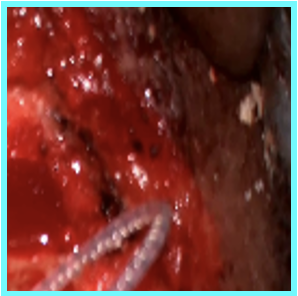}
        \end{minipage}
    \end{minipage}
    \hfill
    \begin{minipage}[t]{0.24\textwidth}
        \begin{minipage}{0.50\textwidth}
            \includegraphics[width=\textwidth]{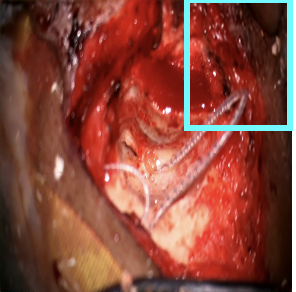}
        \end{minipage}\hspace{-0.3em}
        \begin{minipage}{0.50\textwidth}
            \includegraphics[width=\textwidth]{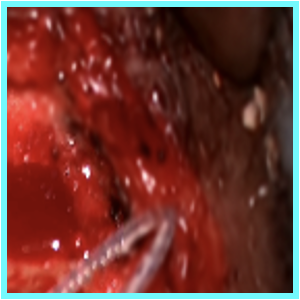}
        \end{minipage}
    \end{minipage}
    \hfill
    \begin{minipage}[t]{0.24\textwidth}
        \begin{minipage}{0.50\textwidth}
            \includegraphics[width=\textwidth]{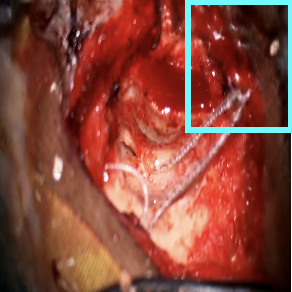}
        \end{minipage}\hspace{-0.3em}
        \begin{minipage}{0.50\textwidth}
            \includegraphics[width=\textwidth]{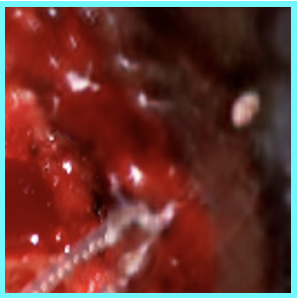}
        \end{minipage}
    \end{minipage}
    \hfill
    \begin{minipage}[t]{0.24\textwidth}
        \begin{minipage}{0.50\textwidth}
            \includegraphics[width=\textwidth]{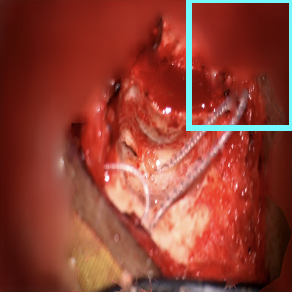}
        \end{minipage}\hspace{-0.3em}
        \begin{minipage}{0.50\textwidth}
            \includegraphics[width=\textwidth]{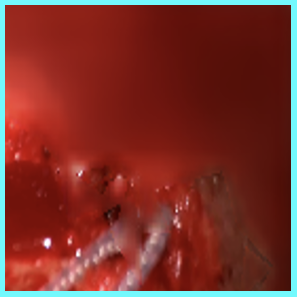}
        \end{minipage}
    \end{minipage}
    \hfill

    \begin{minipage}[t]{0.24\textwidth}
        \begin{minipage}{0.50\textwidth}
            \includegraphics[width=\textwidth]{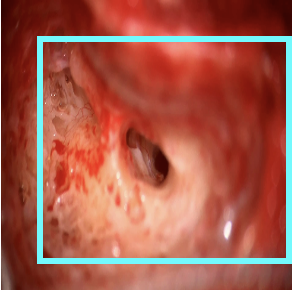}
        \end{minipage}\hspace{-0.3em}
        \begin{minipage}{0.50\textwidth}
            \includegraphics[width=\textwidth]{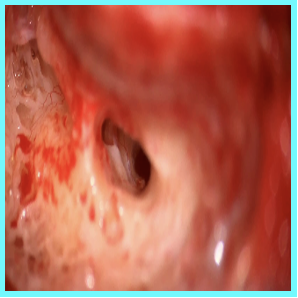}
        \end{minipage}
    \end{minipage}
    \hfill
    \begin{minipage}[t]{0.24\textwidth}
        \begin{minipage}{0.50\textwidth}
            \includegraphics[width=\textwidth]{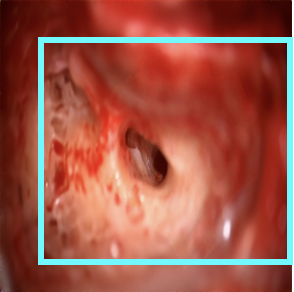}
        \end{minipage}\hspace{-0.3em}
        \begin{minipage}{0.50\textwidth}
            \includegraphics[width=\textwidth]{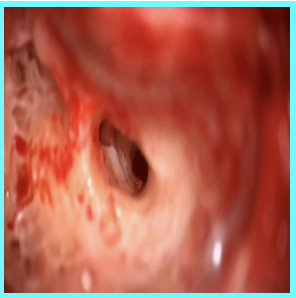}
        \end{minipage}
    \end{minipage}
    \hfill
    \begin{minipage}[t]{0.24\textwidth}
        \begin{minipage}{0.50\textwidth}
            \includegraphics[width=\textwidth]{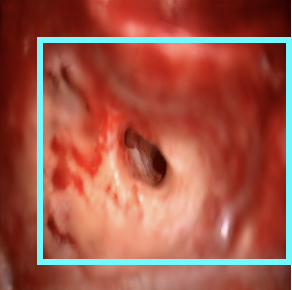}
        \end{minipage}\hspace{-0.3em}
        \begin{minipage}{0.50\textwidth}
            \includegraphics[width=\textwidth]{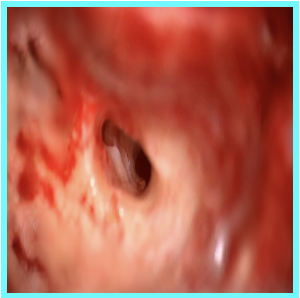}
        \end{minipage}
    \end{minipage}
    \hfill
    \begin{minipage}[t]{0.24\textwidth}
        \begin{minipage}{0.50\textwidth}
            \includegraphics[width=\textwidth]{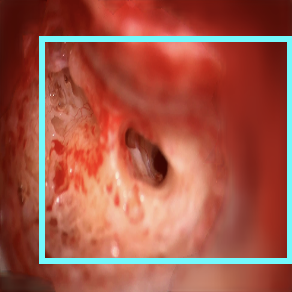}
        \end{minipage}\hspace{-0.3em}
        \begin{minipage}{0.50\textwidth}
            \includegraphics[width=\textwidth]{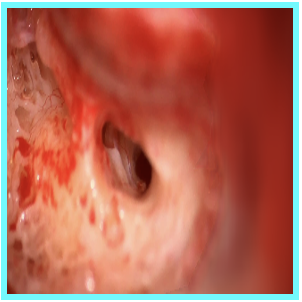}
        \end{minipage}
    \end{minipage}
    \hfill

    \begin{minipage}[t]{0.24\textwidth}
        \begin{minipage}{0.50\textwidth}
            \includegraphics[width=\textwidth]{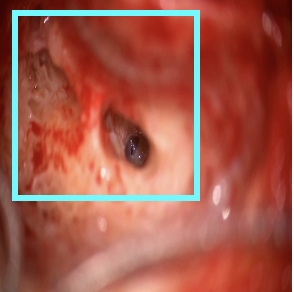}
        \end{minipage}\hspace{-0.3em}
        \begin{minipage}{0.50\textwidth}
            \includegraphics[width=\textwidth]{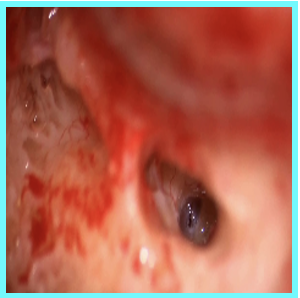}
        \end{minipage}
    \end{minipage}
    \hfill
    \begin{minipage}[t]{0.24\textwidth}
        \begin{minipage}{0.50\textwidth}
            \includegraphics[width=\textwidth]{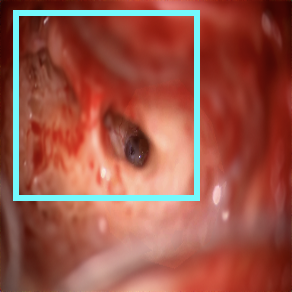}
        \end{minipage}\hspace{-0.3em}
        \begin{minipage}{0.50\textwidth}
            \includegraphics[width=\textwidth]{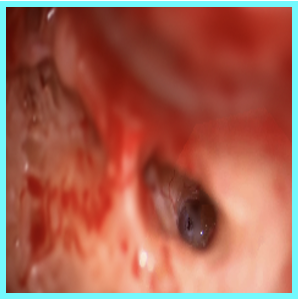}
        \end{minipage}
    \end{minipage}
    \hfill
    \begin{minipage}[t]{0.24\textwidth}
        \begin{minipage}{0.50\textwidth}
            \includegraphics[width=\textwidth]{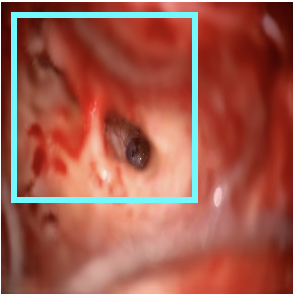}
        \end{minipage}\hspace{-0.3em}
        \begin{minipage}{0.50\textwidth}
            \includegraphics[width=\textwidth]{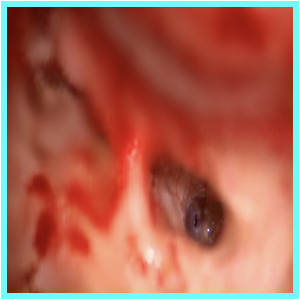}
        \end{minipage}
    \end{minipage}
    \hfill
    \begin{minipage}[t]{0.24\textwidth}
        \begin{minipage}{0.50\textwidth}
            \includegraphics[width=\textwidth]{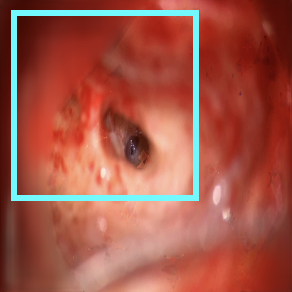}
        \end{minipage}\hspace{-0.3em}
        \begin{minipage}{0.50\textwidth}
            \includegraphics[width=\textwidth]{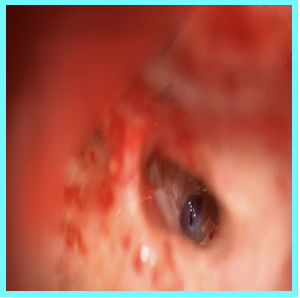}
        \end{minipage}
    \end{minipage}
    \hfill

    \begin{minipage}[t]{0.24\textwidth}
        \begin{minipage}{0.50\textwidth}
            \includegraphics[width=\textwidth]{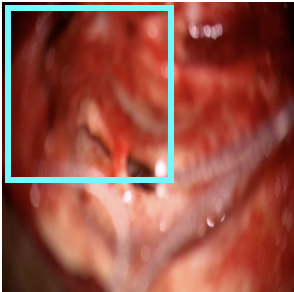}
        \end{minipage}\hspace{-0.3em}
        \begin{minipage}{0.50\textwidth}
            \includegraphics[width=\textwidth]{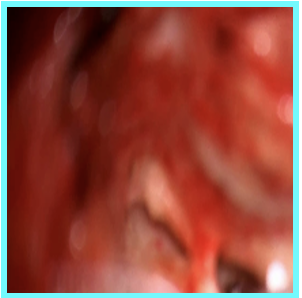}
        \end{minipage}
    \end{minipage}
    \hfill
    \begin{minipage}[t]{0.24\textwidth}
        \begin{minipage}{0.50\textwidth}
            \includegraphics[width=\textwidth]{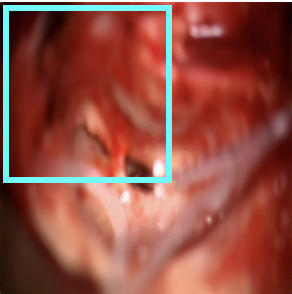}
        \end{minipage}\hspace{-0.3em}
        \begin{minipage}{0.50\textwidth}
            \includegraphics[width=\textwidth]{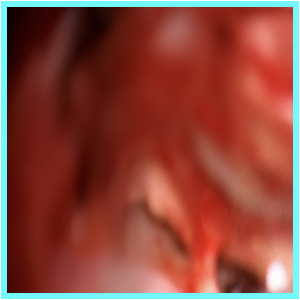}
        \end{minipage}
    \end{minipage}
    \hfill
    \begin{minipage}[t]{0.24\textwidth}
        \begin{minipage}{0.50\textwidth}
            \includegraphics[width=\textwidth]{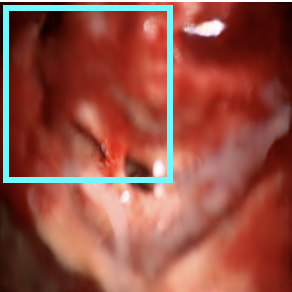}
        \end{minipage}\hspace{-0.3em}
        \begin{minipage}{0.50\textwidth}
            \includegraphics[width=\textwidth]{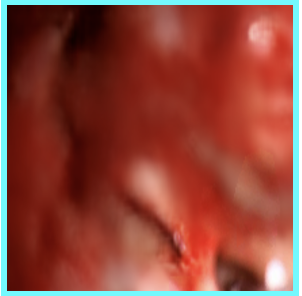}
        \end{minipage}
    \end{minipage}
    \hfill
    \begin{minipage}[t]{0.24\textwidth}
        \begin{minipage}{0.50\textwidth}
            \includegraphics[width=\textwidth]{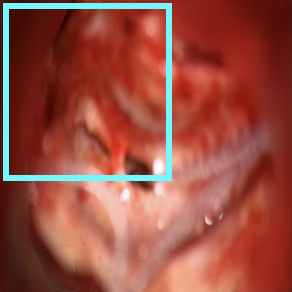}
        \end{minipage}\hspace{-0.3em}
        \begin{minipage}{0.50\textwidth}
            \includegraphics[width=\textwidth]{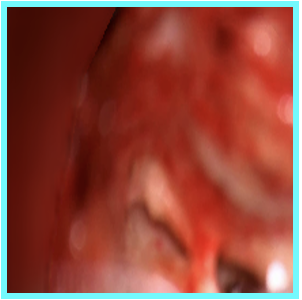}
        \end{minipage}
    \end{minipage}
    \hfill
    
    \begin{minipage}[t]{0.24\textwidth}
        \begin{minipage}{0.50\textwidth}
            \includegraphics[width=\textwidth]{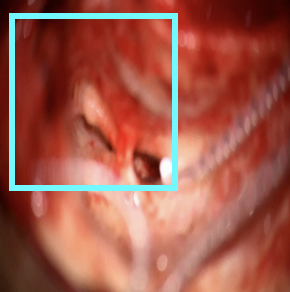}
        \end{minipage}\hspace{-0.3em}
        \begin{minipage}{0.50\textwidth}
            \includegraphics[width=\textwidth]{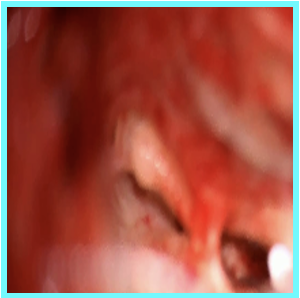}
        \end{minipage}

        \centering
        \vspace{0.3em}
        \footnotesize{GT}
    \end{minipage}
    \hfill
    \begin{minipage}[t]{0.24\textwidth}
        \begin{minipage}{0.50\textwidth}
            \includegraphics[width=\textwidth]{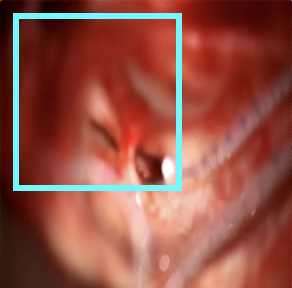}
        \end{minipage}\hspace{-0.3em}
        \begin{minipage}{0.50\textwidth}
            \includegraphics[width=\textwidth]{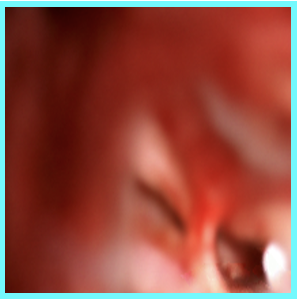}
        \end{minipage}
        
        \centering
        \vspace{0.3em}
        \footnotesize{SSDD-GAN}
    \end{minipage}
    \hfill
    \begin{minipage}[t]{0.24\textwidth}
        \begin{minipage}{0.50\textwidth}
            \includegraphics[width=\textwidth]{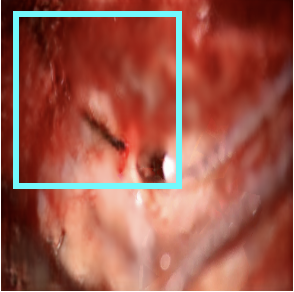}
        \end{minipage}\hspace{-0.3em}
        \begin{minipage}{0.50\textwidth}
            \includegraphics[width=\textwidth]{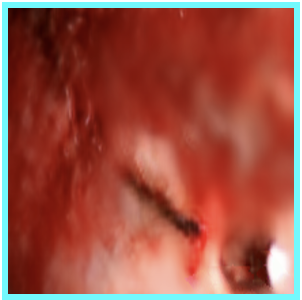}
        \end{minipage}
        
        \centering
        \vspace{0.3em}
        \footnotesize{DeepFillv2}
    \end{minipage}
    \hfill
    \begin{minipage}[t]{0.24\textwidth}
        \begin{minipage}{0.50\textwidth}
            \includegraphics[width=\textwidth]{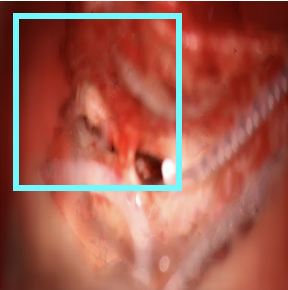}
        \end{minipage}\hspace{-0.3em}
        \begin{minipage}{0.50\textwidth}
            \includegraphics[width=\textwidth]{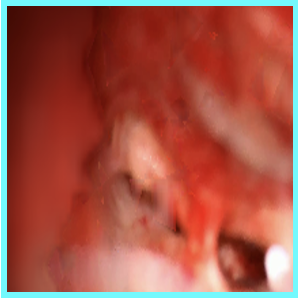}
        \end{minipage}
        
        \centering
        \vspace{0.3em}
        \footnotesize{Pix2Pix}
    \end{minipage}
    \hfill

    \caption{\textbf{Qualitative Comparison.} All images have been resized to 256×256 pixels, with cyan bounding boxes highlighting detailed regions. Each column presents the results alongside their corresponding zoomed-in sections in cyan boxes.}
    \label{fig:figure6}
\end{figure*}

Figure \ref{fig:figure6} presents more qualitative comparisons of various samples with different mask ratios, evaluated using top-performing methods as ranked by the SSIM scores listed in Table~\ref{Tab:quantitative}.
\vspace{-4em}
\end{document}